\documentclass[letterpaper,twocolumn,10pt]{article}
\usepackage{usenix/usenix-2020-09}

\usepackage{amsfonts}
\usepackage{amsmath}
\usepackage{amssymb}
\usepackage{mathrsfs}
\usepackage{tikz}

\usepackage[ruled,linesnumbered]{algorithm2e}
\usepackage{booktabs} 
\usepackage{multirow}
\usepackage{subcaption}

\usepackage{filecontents}
\usepackage{mathtools}

\begin{document}

\date{}

\title{Constrained Adaptive Attacks: Realistic Evaluation of Adversarial Examples and Robust Training of Deep Neural Networks for Tabular Data}

\author{
{\rm Thibault Simonetto}\\
University of Luxembourg
\and
{\rm Salah Ghamizi}\\
University of Luxembourg / RIKEN AIP
\and
{\rm Antoine Desjardins}\\
University of Luxembourg
\and
{\rm Maxime Cordy }\\
University of Luxembourg
\and
{\rm Yves Le Traon}\\
University of Luxembourg
} 

\maketitle

\begin{abstract}
State-of-the-art deep learning models for tabular data
have recently achieved acceptable performance to be deployed in industrial settings. 
However, the robustness of these models remains scarcely explored. 
Contrary to computer vision, there is to date no realistic protocol to properly evaluate the adversarial robustness of deep tabular models due to intrinsic properties of tabular data such as categorical features, immutability, and feature relationship constraints.
To fill this gap, we propose CAA, the first efficient evasion attack for constrained tabular deep learning models. CAA is an iterative parameter-free attack that combines gradient and search attacks to generate adversarial examples under constraints.  
We leverage CAA to build a benchmark of deep tabular models across three popular use cases: credit scoring, phishing and botnet attacks detection. Our benchmark supports ten threat models with increasing capabilities of the attacker, and reflects real-world attack scenarios for each use case.
Overall, our results demonstrate how domain knowledge, adversarial training, and attack budgets impact the robustness assessment of deep tabular models and provide security practitioners with a set of recommendations to improve the robustness of deep tabular models against various evasion attack scenarios.  

\end{abstract}

\section{Introduction}
Evasion attacks are the process of carefully crafting inputs designed to force a machine learning (ML) model to output a wrong decision. These inputs - named adversarial examples - are required to be close to legitimate inputs. 
Robustness to adversarial examples is a problem of growing concern among the secure ML community, with over 10,000 publications on the subject since 2014~\cite{carlini2019evaluating}. 

These growing concerns have since transpired to the public debate and lawmakers are investigating multiple regulations to mitigate these threats. The EU Artificial Intelligence Act~\cite{siegmann2022brussels} has spearheaded the regulations on robust and trustworthy ML and proposed multiple mitigation strategies, including auditing the robustness and security of ML systems in critical fields such as financial machine learning and healthcare.

These domains, however, heavily rely on tabular machine learning models~\cite{Shwartz-Ziv2021,borisov2021robust} that pose unique challenges to train and optimize. 
First, tabular models rely on heterogeneous features such as categorical and discrete features that remain challenging to deep learning models~\cite{hancock2020survey}. Next, tabular data exhibit complex relationships and constraints involving multiple features. The satisfaction of these constraints can be a non-convex or even non-differentiable problem that gradient-based optimizations incorrectly handle~\cite{simonetto2021unified}. Finally, tabular machine learning in production can involve specific feature engineering, that is "secret" and inaccessible to a third party. For example, in credit scoring applications, Ghamizi et al.~\cite{ghamizi2020search} pointed out that while the end-user has an impact on some features of the deep learning model and could perturb them, many features are processed and extended with additional domain knowledge before being fed to the deep learning model.

Research in adversarial robustness for tabular machine learning in general (and tabular deep learning in particular) is still in its infancy and does not yet properly handle these unique properties. The current state of robust tabular deep learning is a stark contrast to the abundant literature on adversarial robustness in computer vision~\cite{survey_advs} and natural language processing tasks~\cite{dyrmishi2023humans}.

We hypothesize that the tabular machine learning community faces three main challenges for research in adversarial tabular deep learning to flourish: (1) a reliable evaluation protocol of adversarial robustness tailored to the specificities of tabular machine learning, (2) an efficient and generic evasion attack for tabular deep learning, and (3) a public benchmark studying realistic threat models with a collection of curated datasets and pre-trained models at different levels of robustness.

Our work addresses the first challenge and the related research question:
\begin{description}
\item \emph{RQ1: How to realistically assess the robustness of tabular deep learning models?}
\end{description}
We answer this question with an exhaustive protocol of ten evaluation scenarios focused on the attacker's capabilities and knowledge of the target tabular deep learning model. We argue that access to domain knowledge and training distribution are critical components for successful evasion attacks.

\begin{figure}
    \centering
    \includegraphics[width=.9\linewidth]{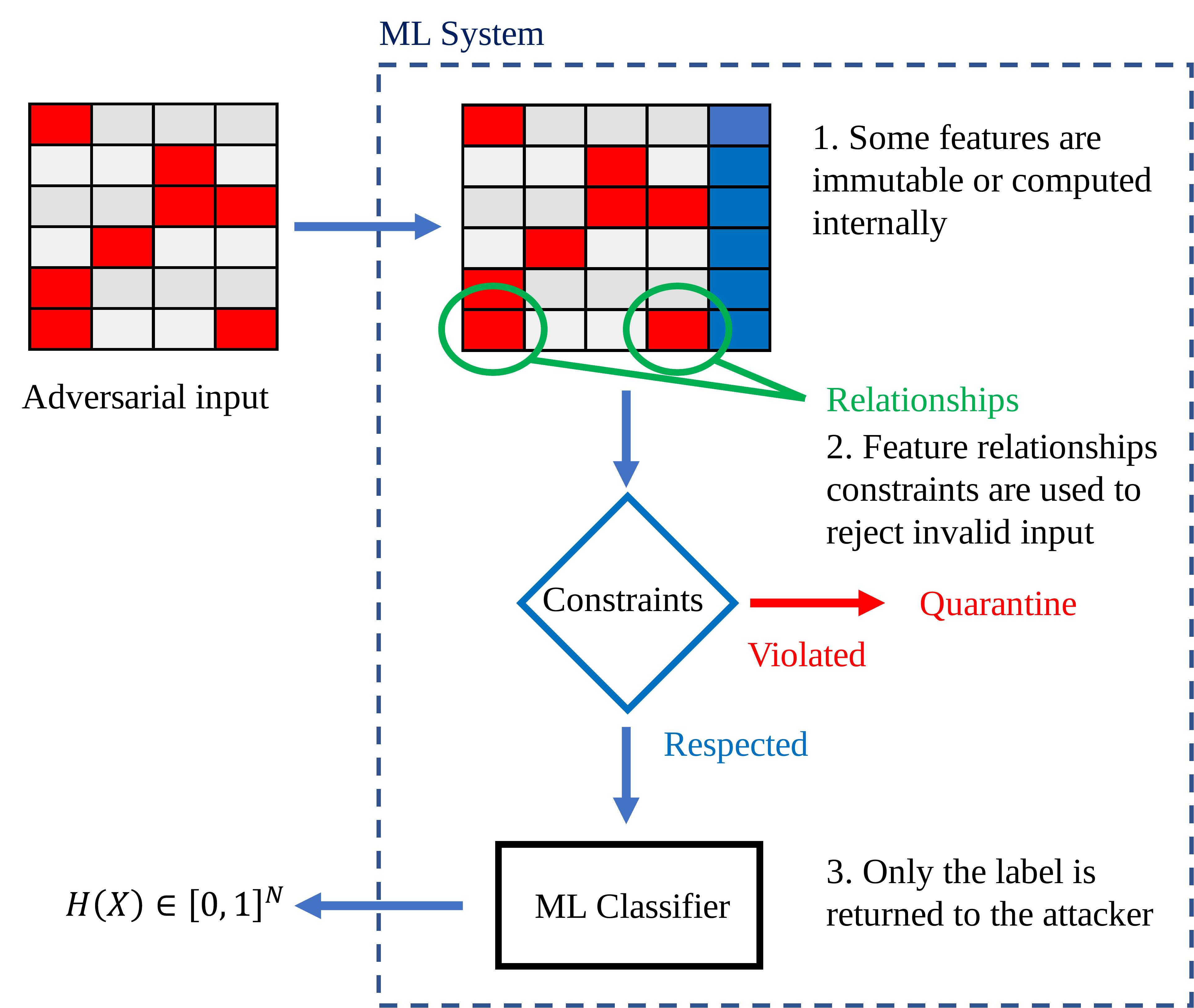}
    \caption{The main challenges for adversarial attacks in Tabular Machine Learning: When an adversary perturbs some features (red), it may not be aware of the new features that are computed internally and added (blue), or the relationships between features (green). If the monitoring system detects a constraint violation, the input is quarantined and a 1 (rejection) is returned to the attacker.}
    \label{fig:challenges}
\end{figure}

Next, we tackle the second challenge:
\begin{description}
\item \emph{RQ2: How to effectively and efficiently attack tabular deep learning models?}
\end{description}
We provide a new effective gradient and parameter-free attack, Constrained Adaptive Attack (CAA). We exemplify in Figure \ref{fig:challenges} the challenges that CAA overcomes to effectively attack tabular deep learning models: Our attack is capable of correctly handling categorical features, feature relationships, and non-differential constraints with an iterative process that maximizes the error of the model while minimizing the constraints violations. 

CAA combines two gradient-based attacks CPGD (Constrained Projected Descent) and CAPGD (Constrained Adaptive Project Descent), and a search-based attack MOEVA \cite{simonetto2021unified} to efficiently adapt the search strategy to tabular datasets of increasing complexity.  
In our empirical study, we demonstrate that CAA is the best candidate for a standardized benchmark and can reliably be used to compare and rank architectures across multiple datasets. 

Our work addresses the last challenge with an extensive benchmark of the robustness of tabular machine learning models in ten realistic attack threat models.
Our threat models consider different levels of knowledge of the model architecture, of the training data, and of the domain of application.

This benchmark addresses our third research question:
\begin{description}
    \item \emph{RQ3: Which defender’s capabilities decrease the success rate of adversarial examples?}
\end{description}

and its counterpart:
\begin{description}
    \item \emph{RQ4:  Which attacker’s capabilities increase the success rate of adversarial examples?}
\end{description}

We draw from these questions a set of recommendations and takeaways for practitioners on both sides and suggest future research directions that remain critical to explore for reliable deployment of trustworthy deep learning models for tabular data.

We benchmark three main families of tabular deep learning model~\cite{borisov2021robust}: Encoding, Transformers, and Regularization models. We evaluate these families with standard and adversarial training on three popular constrained datasets: a phishing detection dataset (URL \cite{hannousse2021towards}), a credit scoring dataset (LCLD \cite{lcld}), and a botnet detection dataset (CTU \cite{chernikova2022fence}). For each dataset, we designed a set of constraints with increasing complexity and evaluated the effectiveness of four adversarial attacks across the three architectures.

To the best of our knowledge, our work is the first to evaluate the robustness of deep learning models for tabular data in realistic scenarios, by providing new effective gradient and parameter-free attacks, and an extensive assessment of the robustness of state-of-the-art tabular deep learning models in realistic threat models and scenarios.

\paragraph{Contributions.} Our contributions can be summarized as follows:
\begin{enumerate}
\item We study ten realistic scenarios and build the first benchmark of adversarial robustness of binary tabular machine learning with three state-of-the-art models and three constrained tabular machine learning datasets.
\item We propose a new efficient and effective evasion attack (CAA) that takes into account the specificities of tabular machine learning.
\item Using our benchmark, we demonstrate that our attack CAA outperforms the best tabular machine learning attacks and is up to five times more efficient. 
\item We derive from our extensive benchmark major takeaways for an attacker and a defender on the impact of model architectures, defense assumptions, and adversarial training effectiveness.

\end{enumerate}

\section{Related work}

\subsection{Tabular Deep Learning}

Tabular data remains the most commonly used form of data~\cite{Shwartz-Ziv2021}, especially in critical applications such as medical diagnosis~\cite{ulmer2020trust, somani2021deep, borisov2021robust}, financial applications~\cite{ghamizi2020search,clements2020sequential,cartella2021adversarial}, user recommendation systems~\cite{zhang2019deep}, customer churn prediction~\cite{ahmed2017survey, tang2020customer}, cybersecurity~\cite{chernikova2019fence,castro2019aimed,aghakhani2020malware,rosenberg2020query}, and more. Improving the performance and robustness of tabular machine learning models for these applications is becoming critical as more ML-based solutions are cleared to be deployed in critical settings.

Borisov et al.~\cite{borisov2021deep} showed that deep neural networks tend to yield less favorable results in handling tabular data when compared to more traditional machine learning methods, such as tree-based approaches. They suggested four main reasons specific to tabular data, namely \textbf{low-quality} training data,  complex irregular \textbf{spatial dependencies} between features, \textbf{sensitivity to preprocessing}, and \textbf{imbalanced importance} of features. To overcome these challenges, the tabular ML community proposed various optimizations that can be sorted across 3 families: \textbf{data transformation methods}, such as VIME~\cite{yoon2020vime}, that prepend some encoding operations to help deep neural networks to better extract the relevant information. Next, are methods with \textbf{specialized architectures}, such as TabTransformer~\cite{huang2020tabtransformer}, that are designed specifically for heterogeneous tabular data. Finally, \textbf{regularization models} propose novel loss functions and training processes. Among the most popular methods in this category is the regularization learning network (RLN) proposed by Shavitt et al.~\cite{shavitt2018regularization}. 

In our study, we evaluate one representative method of each category based on their test performance in the survey by Borisov et al.~\cite{borisov2021deep}. 

\subsection{Realistic Adversarial Examples}

Initially, adversarial machine learning research focused on demonstrating the vulnerability of ML models in the worst-case scenarios, the white-box setting \cite{biggio2013evasion,papernot2017practical}. Recently, research has shifted its focus towards areas where the process of feature perturbation is more constrained, with black-box attack scenarios \cite{andriushchenko2020square,rosenberg2020query} and problem space attacks\cite{pierazzi2020intriguing}. In these scenarios, adversaries are required to manipulate objects within the problem space, without having precise knowledge of how these modifications will impact the feature space. This challenge is referred to as the "inverse feature-mapping" problem, as documented in various works such as Dyrmishi et al.'s\cite{dyrmishi2022empirical}, Pierazzi et al.'s~\cite{pierazzi2020intriguing}, Konrad's~\cite{quiring2019misleading}, and Biggio's~\cite{Biggio_2013}. In the work by Dyrmishi et al.\cite{dyrmishi2022empirical} in particular, the authors explored the effectiveness of different adversarial attacks on natural language processing tasks, malware classification, and botnet attack detection. They demonstrated the high cost of running full problem space attacks and suggested that feature space attacks under constraint satisfaction can be effective alternatives to assess the robustness of tabular machine learning in realistic scenarios at a fraction of the cost of problem space attacks.

Our work follows this hypothesis and focuses on constrained feature-space attacks to realistically assess the robustness of deep tabular learning models.

\subsection{Black-box attacks}

Black-box attacks refer to attacks that do not have access to the internals of the target model. 
Two main mechanisms are leveraged to generate the attacks. First, attacks can be designed to maximize the transferability from a source model (that is attacked in a white-box setting) to the target model. Szegedy~\cite{szegedy2013intriguing} was first to demonstrate that an adversarial example generated against one model with a gradient-based attack can be transferred to another model in an untargeted setting with a relatively high success rate. Techniques that rely on gradient-based transferability can be broken down into several components~\cite{Zhao2022TowardsAttacks}, including model augmentation, data augmentation, attack optimizers, and feature-based attacks.

Our research tackles multiple scenarios with transfer attacks, however, our aim is not to benchmark the plethora of techniques that improve the transferability of attacks in a tabular setting. This work is an orthogonal endeavor, and we focus on the transferability of the strongest white-box attack of the literature, AutoAttack~\cite{croce2020reliable}. 

A second line of research on black-box attacks focuses on query-based attacks, where the target model is queried with a limited number of inputs, and its outputs are used to generate better adversarial examples. Early work of this family was the ZOO attack ~\cite{chen2017zoo}, where the queries are used to approximate the gradient of the target model. The successive improvements focused on reducing the number of queries (SimBA-DCT~\cite{guo2019simple}, TREMBA~\cite{huang2019black}, BASES~\cite{cai2022blackbox}). 

Similarly to transfer attacks, we do not aim to benchmark the large literature of query-based attacks when transposed to the tabular setting, but focus our work on one representative of this family: MOEVA~\cite{simonetto2021unified}. This attack was designed for the same datasets of our study and has been shown to particularly fit the complex case of tabular machine learning with domain constraints.
\section{Realistic Robustness Evaluation for Tabular ML}

Our first contribution is to propose an exhaustive evaluation protocol for the adversarial robustness evaluation of tabular deep learning models.

We first consider the problem of generating adversarial examples under domain constraints as a cornerstone to any realistic evaluation of tabular deep learning model's robustness. We formulate this problem as a multi-objective search and extend the classification definition of adversarial examples to constrained adversarial examples in section \ref{sec:pb}.

Next, we propose ten evaluation scenarios that gradually restrict the capabilities of the attacker and allow practitioners to fine-tune their evaluation to the actual security requirements of the system. We present these scenarios in section \ref{sec:threat_models} and evaluate each of them in our empirical study (section \ref{sec:results}).

\subsection{Problem Formulation}\label{sec:pb}

We formulate below the problem of generating adversarial examples that respect domain constraints. We extend the constraint framework defined in~\cite{simonetto2021unified}. 

The problem of generating adversarial examples to evaluate the robustness of ML systems has been extensively studied in the domain of computer vision.
The problem is the following. Let $x \in \mathbb{R}^d$ be an input point and $y \in \{1, \dots,  C\}$ be its correct label.
Given a classifier $h: \mathbb{R}^d \rightarrow \mathbb{R}^C$ we define a successful adversarial example with respect to the allowed perturbation set $\Delta \subseteq \mathbb{R}^d$ as a vector $\delta \in \mathbb{R}^d$ such that 
\begin{equation}
\label{eq:adv-pb}
    argmax_{c \in \{1, \dots, C\}} h(x+\delta) \neq y \,\text{ and }\, \delta \in \Delta
\end{equation}

For image classification the set $\Delta$ of allowed perturbation is chosen as the $l_p$-perturbations, i.e. $\Delta_p = \{ \delta \in \mathbb{R}^d, ||\delta||_p \leq \epsilon\}$, where $\epsilon$ represents the maximum allowed perturbation size.

By bounding the maximum perturbation allowed for the attack, we aim at preserving the semantic of the image: we want $x$ and $x+\delta$ to be perceived similarly by then human such that we can suppose $x + \delta$ also have $y$ as true label.
This also guarantees that the image is still somewhat meaningful, that is $x+\delta$ is not simply random noise.
The only additional constraints, except for the perturbation size, is that each feature $x_i \in \mathbb{d}, i \in d$ is a value between 0 and 1, i.e. $x_i \in [0, 1]$.
These constraints are straightforward to satisfy using projection \cite{madry2017towards}.
    
Tabular data are by nature different from image data.
Tabular data typically represents objects of the considered application domain (e.g. botnet \cite{chernikova2022fence}, financial transaction \cite{ghamizi2020search}). We assume the existence of feature mapping $\varphi: Z \rightarrow \mathcal{X} \subseteq \mathbb{R}^d$ that maps objects of the input space $Z$ to a $d$-dimensional feature space $\mathcal{X}$.
Each object $z \in Z$ respects some natural condition to be valid. In the feature space, these conditions translate into a set of constraints over the feature values, which we denote $\Omega$.
By construction, any feature vector $x$ generated from a real-world object $z$ satisfies all constraints $\omega \in \Omega$. We denote $\mathcal{X}_\Omega$ the subset of feature vectors in $\mathcal{X}$ that satisfies the set of constraints $\Omega$.

The objective is to generate adversarial examples that are realistic, therefore generated examples must satisfy the constraints domain $x + \delta \in \mathcal{X}_\Omega$.
From Equation~\ref{eq:adv-pb}, that is equivalent to $\Delta_p = \{\delta \in \mathbb{R}, ||\delta_p|| \leq \epsilon \land  x+\delta \in \mathcal{X}_\Omega \}$.
Note that we keep the bound on the perturbation size and suppose that $x + \delta$ also has $y$ as ground truth.

In our evaluation, we support the four types of constraints introduced by Simonetto et al.~\cite{simonetto2021unified}:
\paragraph{Mutability constraints} define what feature can be modified by an attacker.
\paragraph{Boundary constraints} define the upper bound and lower bound of each feature.
\paragraph{Types constraints} define the type of each feature as continuous numerical, discrete numerical, or categorical.
\paragraph{Feature relation constraints} capture the relation between two or more features using a constraints grammar.

We use the constraints grammar from  \cite{simonetto2021unified}:
 \begin{align}
  \omega &\coloneqq \omega_1 \land \omega_2 \mid \omega_1 \lor \omega_2 \mid \psi_1 \succeq \psi_2 \mid f \in \{\psi_1 \dots \psi_k\}\\
    \psi &\coloneqq c \mid f \mid \psi_1 \oplus \psi_2 \mid x_i
\end{align}

 where $f \in F$, $c$ is a constant real value, $\omega, \omega_1, \omega_2$ are constraint formulae, $\succeq \in \{<, \leq, =, \neq, \geq, >\}$, $\psi, \psi_1, \dots, \psi_k$ are numeric expressions, $\oplus \in \{+, -, *, /\}$, and $x_i$ is the value of the $i$-th feature of the original input $x$. 
For instance, for our credit use case, the number of open accounts and the number of total accounts are constrained by the relation
\begin{align}
    open\_accounts \leq total\_accounts
\end{align}
where $open\_accounts$ and $total\_accounts$ are two features used by the prediction model.

We express each relationship constraints $\omega_i$ as a penalty function $penalty(x, \omega_i)$ such that $x$ satisfies $\omega_i$ if and only if $penalty(x,\omega_i) = 0$.
This penalty function represents how far an example $x$ is from satisfying the constraint $\omega_i$.
Table \ref{tab:grammar-minimisation} shows how each constraint translates to a penalty function. 
We use the penalty function of the conjunction of the set of domain constraints to guide the search toward constrained adversarial examples.

\begin{table}

  \centering
  \caption{From constraint formulae to penalty functions. $\tau$ is an infinitesimal value.}
  \label{tab:grammar-minimisation}
  \small
 
  \begin{tabular}{llll}
    ID & Constraints formulae     & Penalty function   \\
    \hline
    $\land$ &$\omega_1 \land \omega_2$ & $\omega_1 + \omega_2$ \\
     $\lor$ &$\omega_1 \lor \omega_2$ & $ \min(\omega_1, \omega_2)$ \\
    $\in$ &$\psi \in \Psi = \{\psi_1, \dots \psi_k\}$  & $\min(\{\psi_i \in \Psi: \mid \psi - \psi_i \mid\})$ \\
    $\leq$ &$\psi_1 \leq \psi_2$  & $max(0, \psi_1 - \psi_2)$   \\
    $<$ &$\psi_1 < \psi_2$ & $max(0, \psi_1 - \psi_2 + \tau)$   \\
    $=$ &$\psi_1 = \psi_2$ & $\mid \psi_1 - \psi_2 \mid$ \\
   
  \end{tabular}
  
\end{table}

\subsection{Realistic Threat Models}
\label{sec:threat_models}
We claim that there is a need in tabular ML evasion attacks for realistic threat models that are tailored to the requirements of the deep learning system. In our setting, the adversary's \textbf{goal} is to flip the label of a binary classification from the legitimate label to the adversary label. 
We investigate for the attacker three \textbf{dimensions of capabilities} of the target system, and we summarize our scenarios in Table \ref{tab:scenarios}:

\textbf{Model access}. When the attacker has full access to the model in a white-box manner, gradient-based attacks are possible. Otherwise, the attacker can only query the target model for logits, or needs to design a surrogate model to transfer the attack in the black-box setting. For each setting, we evaluate both the cases where the models are trained using standard training, and when models are trained with adversarial training.

\textbf{Domain knowledge} defines whether the attacker is aware of the domain-specific properties of the system, such as the relationships between the features and the constraints of the system. We assume when the dataset is accessible that the attacker has access to the boundaries and type of features involved in the machine learning model.

\textbf{Dataset knowledge}. In most scenarios, the attacker is assumed to have access to the training distribution of the target model, but we also investigate the extreme scenarios where the attacker has only access to a subset of the training set (e.g., 10\%) or only access to a dataset of the same distribution as the training set. 

We build our ten evaluation scenarios by successively reducing the knowledge and capability of the attacker, from scenario A1 where the attacker has full access and knowledge of the target (the scenario commonly evaluated in the literature) to scenario E2 where the attacker has no access to the model, nor the training set and has no domain-knowledge about the target system. These scenarios can be seen as respectively the worst- and best-case scenarios for the defender. 

Each scenario A, B, C, D, and E has two variants. The first one (A1, ..., E1) assumes that the attacker is aware of the domain-specific properties of the system (the scenarios in Table \ref{tab:scenarios} where \textbf{Domain knowledge} is True). The second variant (A2, ..., E2) assumes that this knowledge is unavailable to the attacker. The five base scenarios are:  

\paragraph{A - Full White-box Attacks} In this scenario, the attacker has directly access to the target model and its weights. The attacker also has access to the full training and test sets of the model and launches its attacks from genuine examples of the test set.
The attacker can run as many inference steps as needed and can collect the loss and gradient of each inference step.

\paragraph{B - Query-based Attacks}In this scenario, the attacker can query the model and collect the binary output. The attacker has access to the full training and test sets of the model and launches its attacks from genuine examples of the test set.

\paragraph{C - Transfer-based Attacks}This scenario is also referred to as "surrogate-based attacks" because the attacker has full access to the training set of the target, but no knowledge of the hyperparameters or the architecture used in training the target model.

\paragraph{D - Transfer-based Attacks}This scenario is identical to scenario C, except that the attacker only manages to collect a limited subset of the training set of the target model. 

\paragraph{E - Distribution Transfer-based Attacks}This scenario is identical to the scenario C, except that the attacker only manages to collect a set of the same distribution as the training set of the target. It can mimic the use of generative models or the use of datasets from the same task (eg., a public credit scoring dataset of one bank to attack another).

\begin{table}[t]
    \centering
     \caption{The ten scenarios of our study with decreasing attacker knowledge of the domain constraints (Domain), access to model internals and outputs (Model), and access to the training dataset (Dataset). Y: Yes, Q: Query-only, S: Training dataset subset, D: Same distribution as training dataset, N: No}
    \label{tab:scenarios}
    \small
    \setlength{\tabcolsep}{4pt}
    \begin{tabular}{c|cccccccccc}
     \toprule
        Scenario    & A1 & A2 & B1 & B2 & C1 & C2 & D1 & D2 & E1 & E2 \\ 
        \midrule
        Domain      & Y & N & Y & N & Y & N & Y & N & Y&  N\\ \midrule
        Model       & Y & Y & Q & Q & N & N & N & N & N & N \\ \midrule
        Dataset     & Y & Y & Y & Y & Y & Y & S & S & D & D\\ \bottomrule
    \end{tabular}
   
\end{table}

\section{Constrained Adversarial Attacks}\label{sec:attacks}

We describe below our adversarial attacks for tabular data, CPGD, CAPGD, and CAA. They combine components tailored for tabular data (custom loss functions, constraint validation, $L_p$ projection on mutable features...) and borrow well-established mechanisms from computer vision (momentum, adaptive learning rate, adaptive restart).


Our main attack, Constrained Adaptive Attack (CAA) is a meta attack that runs successively three sub-attacks with increasing costs, CPGD, CAPGD, and MOEVA \cite{simonetto2021unified}. These three attacks all follow the following procedure.

\paragraph{Perturbation generation.} Each sub-attack computes the adequate perturbation to generate an adversarial example while minimizing the constraint violations. The constraints are handled as loss functions for gradient attacks (CPGD, CAPGD) and as fitness functions for search attacks (MOEVA).

\paragraph{Constraint validation.} The adversarial inputs are validated with a constraint checker. The constraints checker evaluates the penalty function of each constraint  by translating constraints to a numerical function $g: \mathcal{X} \implies \mathbb{R}$ following the transformations in Table \ref{tab:grammar-minimisation}. A constraint is satisfied for an input $x$ if and only if $g(x) = 0$.

\paragraph{Constraint repair.} This step takes as input the original example $x$ the perturbation $\delta$ and a set of constraints. We repair discrete numerical type constraints by rounding the values in the inverse direction of the perturbation to guarantee that the boundary constraints remain satisfied. To avoid circular constraints' violation, only the relation constraints of the form $ f \in F = \psi$ are repaired by setting the value of feature $f$ to $\psi$ hence $\delta_f = \psi - x_f$.

\subsection{The Components of CAA}

\paragraph{CPGD} Constrained Projected Gradient Descent (CPGD) is an adaptation of the well-established PGD attack \cite{madry2017towards} to generate adversarial examples that satisfy domain constraints. 
CPGD integrates the constraints' penalty function as a negative term in the loss that PGD aims to maximize.
Hence, CPGD produces examples iteratively as follows:

\begin{equation}
\begin{aligned}
x^{(k+1)} = R(\Pi_{x+\delta}(x^{(k)} + \eta^{(k)} sgn(\nabla_{x^{(k)}} l(h(x^{(k)}), y) \\ - \sum_{\phi_i} \nabla_{x^{(k)}} penalty(x^{(k)}, \phi_i))))
\end{aligned}
\label{eq:gradient-pgd}
\end{equation}
with  $x^0 = x$, $\theta_h$ the parameters of our predictor $h$, $\Pi$ is a clip function ensuring that $x+\delta$ remains bounded in a sphere around $x$ of a size $\epsilon$ using a norm $p$ and within the lower and upper bound of the constraint, 
and $\nabla_x l$ is the gradient of loss function tailored to our task. 

We compute the gradient using the first-order approximation of the sum of the penalty functions corresponding to our constraints.

The sign function $sgn$ returns a scaled gradient $sgn(x) = x/L_p(x)$ for $p = 2$ norm such that $L_p(sgn(x)) = 1$.

Mutability constraints are enforced by masking the $sgn$ such that the perturbation gradient is null for immutable features.
At each iteration $k$, the repair method $R$ updates the features to fix the broken constraints (if any).

The step $\eta^{(k)} $ is updated following a prior schedule:
$$ \eta^{(k)} = \epsilon \times 10^{-(1+\lfloor k / \lfloor N_{iter} / M \rfloor \rfloor)} $$ 
where $N_{iter}$ is the number of step iterations and $M$ the parameter that controls the lower bound of $\eta^{(k)}$.

\paragraph{CAPGD} The motivation of this new attack, Constrained Adaptive PGD is to introduce the improvements proposed by Dong et al. \cite{dong2018boosting} and Croce et al. \cite{croce2020reliable} to our CPGD attack.
First CAPGD introduces a momentum to the direction of the perturbation\cite{dong2018boosting}.
Let $ \eta^{(k)}$ be the step size at iteration $k$, then the update step is

\begin{equation}
\begin{aligned}
z^{(k+1)}   &= \Pi_{x+\delta}(x^{(k)} + \eta^{(k)} sgn(\nabla_{x^{(k)}} l(h(x^{(k)}), y) \\ 
            &- \sum_{\phi_i} \nabla_{x^{(k)}} penalty(x^{(k)}, \phi_i))) \\
x^{(k+1)}   &= R(\Pi_{x+\delta}(x^{(k)} + \alpha \cdot (z^{(k+1)} - x^{(k)}) \\
            &\phantom{= R(\Pi_{x+\delta}(x^{(k)}} + (1 - \alpha) \cdot (x^{(k)} - x^{(k+1)})
\end{aligned}
\label{eq:gradient-pgd}
\end{equation}
where $\alpha \in [0, 1]$ regulates the influence of the previous update on the current one. 

Second, we introduce a step size adaptation following \cite{croce2020reliable}.
We start with a step size $\eta^{(0)} = 2\epsilon$, and we identify checkpoints $w_0 = 0, w_1, ..., w_n$ at which we decide whether it is necessary to halve the current step size.
We halve the step size if either of two conditions is true 1) since the last checkpoint, the loss increased for at least of fraction of $\rho$ steps, 2) the step has not been reduced at the last checkpoint and the loss is equal to the loss of last checkpoint.

Third, if at checkpoint $w_j$, the step size is halved, we restart from the example that had the highest loss so far, that is $x^{(w_j)} = x_{max}$.
A direct consequence is that the CAPGD algorithm returns the best solution found while CPGD can potentially discard an intermediate solution $x^{(j)}, j < N_{iter}$ with a higher loss than its output $x^{(N_{iter})}$.

\paragraph{MOEVA} Mutli-Objective Evolutionary Adversarial Attack is a search-based attack proposed by Simonetto et al. \cite{simonetto2021unified} using the multi-objective genetic algorithm R-NSGA-III \cite{vesikar2018reference}. The three objectives are constraints violation minimization 
, misclassification maximization using the output probabilty $h(x^{(k)})$ of the model, and minimization of the $p-norm$ distance to clean example $L_p(x^{(k)} - x)$.
This attack is query-based. It only uses the model probability prediction but not the gradient.

\subsection{Constrained Adaptive Attack}

Constrained Adaptive Attack is an attack that iteratively applies CPGD, CAPGD, and MOEVA.
At each iteration, we only apply the next internal attack on the clean examples for which the previous attack failed to generate a constrained adversarial example.
We apply the attacks in increasing order of cost, such that costlier attacks run only for example where cheaper attacks failed.
Therefore, we reduce the overall cost. Algorithm \ref{alg:caa} summarizes the process of CAA.
We start by setting the clean example as adversarial examples (l. 1). The success mask is set to clean examples that are naturally misclassified by the model (l. 2).
For each attack previously described  (l. 3), we run the attack for all examples that are not adversarial yet (l. 4-5).
We create a mask of the attack results that are adversarial, respect the constraints (l. 6), and add such examples to our global list of adversarial (l. 7).
Finally, we update the mask of successful adversarial such that the next attack only runs on unsuccessful ones (l. 8).
We return the list of potentially adversarial examples that respect domain constraints (l. 9).

\begin{figure}
    \centering
    \includegraphics[width=\linewidth]{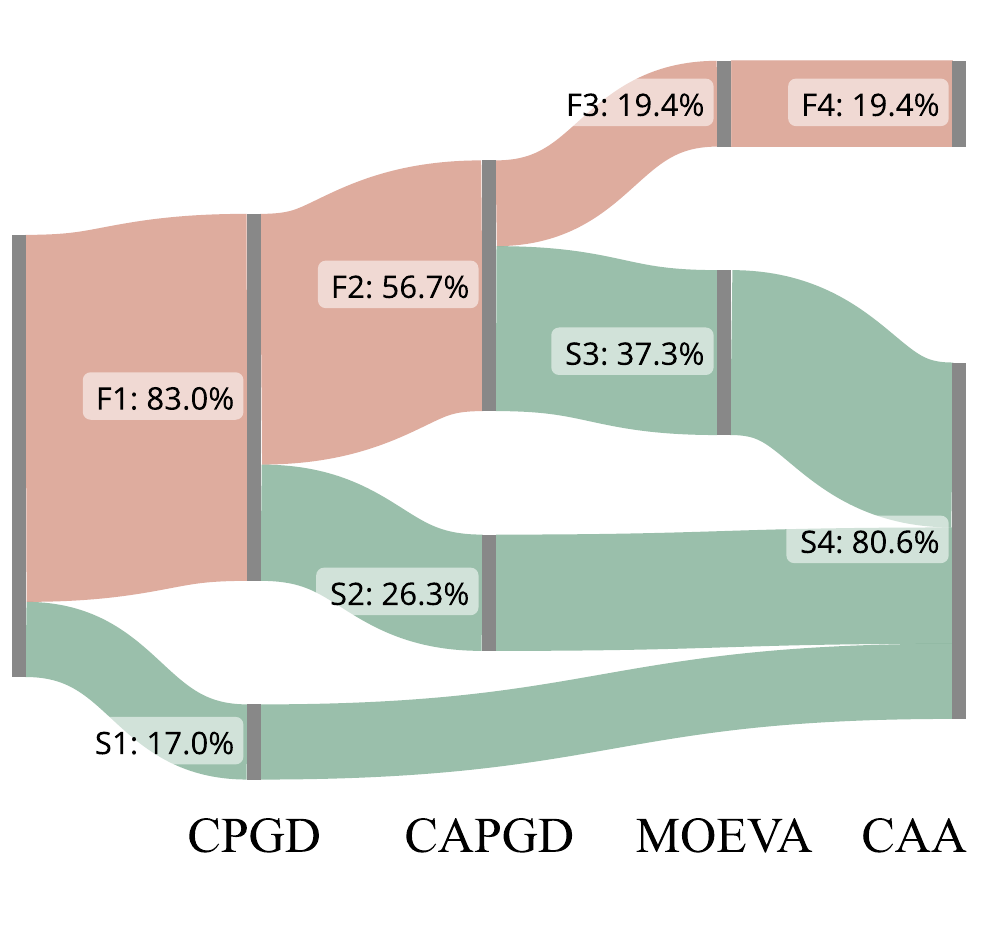}
    \caption{Visualization of the cascading process of CAA for a TabTransformer model trained on LCLD dataset. S1, S2, S3, and S4 in green are respectively the success rates of the attacks CPGD, CAPGD, MOEVA, and CAA. In red are the remaining unsuccessful examples after each attack.}
    \label{fig:caa-sankey}
\end{figure}

The cascading effect of CAA is demonstrated in Figure \ref{fig:caa-sankey}. While CPGD is very efficient, it only generates 17.3\% constrained adversarial examples, CAPGD adds another 15.1\% constrained adversarial examples, and MOEVA 65.5\%. Gradient attacks are computationally more efficient than MOEVA and allow reducing the overall cost of CAA up to 5 folds (see Section \ref{sec:efficient}) 

\begin{algorithm}
    \SetAlgoVlined
    \KwIn{
    $\mathbf{X = (x_i)^N, Y = (y_i)^N}$, a set of N clean examples and their corresponding label\; 
    $\mathbf{Y_0}$, a set of corresponding \; 
    $H$ the model\;
    $ \Omega $ the constraints\;
    }

    \KwOut{$X' = (x'_i)^N$ a set of potentially adversarial example that respect constraints \;}

    $X' = X$ \;
    $success = (H(X') \neq y) \land (X' \in \Omega))$ \;

    \For{$Atck \in \{ CPGD, CAPGD, MOEVA\}$}{
     $X_{step} = X[\neg success]$ \;
     $y_{step} = y[\neg success]$ \;
      $X_{Atck} = Atck(X_{step}, y_{step}, \omega)$ \;
    $success_{Atck} = (H(X_{Atck}) \neq y_{step}) \land (X_{Atck} \in \Omega))$ \;
    
    
    $X'[\neg success][success_{Atck}] = X_{Atck}[success_{Atck}]$ \;
    $success[\neg success] = success_{Atck}$ \;
    }

    \Return{$X'$}
    \caption{Constrained Adaptive Attack (CAA)}
    \label{alg:caa}
\end{algorithm}

Figure \ref{fig:caa-sankey} depicts the execution of CAA. 
We expect CPGD and CAPGD to efficiently generate adversarial example for a significant subset of clean example (here 43.3\%).
MOEVA is then only executed on the remaining 56.7\% of examples, reducing the overall cost of the attack.

\section{Experimental Protocol}

We address the next research questions with an extensive empirical study. We describe below the protocol followed consistently across all the ten threat models.

\subsection{Architectures}

In order to assess our attack, we study how it affects various architectures.  Towards this goal, we study 3 architectures from a recent survey on state-of-the-art models for tabular data, TabSurvey~\cite{borisov2021deep}. The models' best parameters are optimized with Bayesian search for each of our datasets and each model is trained using standard and adversarial~\cite{madry2017towards} training. 

\textbf{TabTransformer} is a transformer-based model~\cite{huang2020tabtransformer}. It uses self-attention to map the categorical features to an interpretable contextual embedding, and the paper claims this embedding improves the robustness of models to noisy inputs. 

\textbf{RLN} or Regularization Learning Networks~\cite{shavitt2018regularization} uses an efficient hyperparameter tuning scheme in order to minimize a counterfactual loss. The authors train a regularization coefficient to weights in the neural network in order to lower the sensitivity and produce very sparse networks.

\textbf{VIME} or Value Imputation for Mask Estimation ~\cite{yoon2020vime} uses self and then semi-supervised learning through deep encoders and predictors.

\subsection{Training}

For each model, we optimize the hyperparameters with cross-validation for 100 iterations then train the models with early stopping for 150 epochs. Suing the same hyperparameters and training protocol, we retrain from scratch with Madry adversarial training \cite{madry2017towards} and $\epsilon=4/255$ with adversarial examples that do not enforce the domain constraints.

\begin{table}[t]
  \centering
  \caption{Clean performance of standardly trained models for the datasets evaluated in the empirical study.}
  \label{tab:data}
  \small
  \begin{tabular}{c|lll}
    \toprule
    Dataset & \multicolumn{3}{|c}{Test performances (AUC)}\\
     & TabTransformer & RLN & VIME\\
    \midrule
    LCLD~\cite{lcld} &  0.717 & 0.719 & 0.714 \\
    CTU-13~\cite{chernikova2022fence}&   0.981 & 0.984 & 0.974 \\
    URL~\cite{hannousse2021towards} &    0.979 & 0.991 & 0.987 \\
    \bottomrule
  \end{tabular}
\end{table}

\subsection{Datasets}

We conduct our study on three established tabular datasets with domain constraints~\cite{simonetto2021unified}.

\textbf{URL} dataset contains legitimate and phishing URLs~\cite{hannousse2021towards}. The features include the number of special characters ("\&", "\$", ","...), the length and port of the URL, the appearance of a brand name, and even the "http" or "https" prefixes. Some features are also computed on the fly from external services such as Google index and PageRank.
This is the simplest dataset of our study because it is a class-balanced dataset with only 14 domain constraints, 7 linear and 7 boolean.

\textbf{Lending Club Loan Data (LCLD)} was originally a Kaggle credit scoring challenge~\cite{lcld}. The inputs are credit requests that can be accepted or rejected according to risk level of non-reimbursement. This dataset has 10 relationship constraints (3 linear and 7 non-linear constraints). While the dataset is moderately unbalanced (80/20), the main challenge remains the non-linearity of the constraints and their complexity.

\textbf{CTU} is a dataset of legitimate and botnet traffic from the CTU University~\cite{chernikova2022fence}. The goal is to predict if the request is made by a human or a botnet. This is a highly unbalanced dataset (99/1), with 360 linear domain constraints across many features. These features are related to network connection protocols and represent the number and type of connections detected in the system. 
The number and relationships between the features make this dataset the most challenging to attack.

We present in Table \ref{tab:data} the test performance achieved by each of the architectures on our three datasets.

\subsection{Adversarial Attacks}

We run all our experiments using an L2 norm of 0.5. For the iterative gradient attacks, we use the default parameters of TorchAttacks~\cite{Kim2020Torchattacks:Attacks}, including ten iterations and a single start. The adaptive step uses a lower bound $M = 7$ for CPGD, and $\alpha = 0.75$ for CAPGD as suggested by Croce et al. \cite{croce2020reliable}.

For search-based attacks, we use 100 generations with a population size of 200 and 100 offsprings. For each of these hyperparameters, we provide ablation studies of their impact. 

For all our experiments, we run the attacks on five seeds and report the mean values across the runs.
We run our experiments on the positive class that represents the critical scenario for our 3 use cases. 
For LCLD it represents a rejected request that would be adversarialy accepted,
for CTU, a malware that would be classified as benign and for URL, a phishing URL that would be classified as legit.

\subsection{Evaluation Metrics}

The models are fine-tuned to maximize cross-validation AUC. This metric is threshold-independent and not affected by the class unbalance of our dataset.

We evaluate the robustness of our model using the adversarial accuracy metric.
We only attack clean examples that are not already misclassified by the model. 
Then, we only consider examples that respect domain constraints to compute the robust accuracy. If an attack generates invalid examples, they are defacto considered unsuccessful and are reverted to their original example (correctly classified). We measure the cost of our attack in execution time.

\section{Results and Discussion}
\label{sec:results}

\subsection{Effectiveness and Efficiency of CAA}
\label{sec:efficient}

\paragraph{CAA is the most efficient constrained attack.}

\begin{table*}
\centering
\caption{Model robust accuracy against different attacks in the white-box scenarios A on different models and datasets. The Clean column corresponds to the accuracy of the model on the subset of clean samples that we attack. A lower robust accuracy means a more effective attack. The lowest robust accuracy is in bold.}
\label{tab:scenario_a_effectiveness}
\small
\begin{tabular}{lll|lllll}
\toprule
    &        & Attack & Clean &                       CPGD &                      CAPGD &                      MOEVA &                        CAA \\
Dataset & Training & Model &          &                            &                            &                            &                            \\
\midrule
\multirow{6}{*}{URL} & \multirow{3}{*}{Standard} & RLN &  $0.944$ &  $0.217$\small{$\pm 0.003$} &  $0.266$\small{$\pm 0.012$} &  $0.236$\small{$\pm 0.005$} &  \textbf{$\textbf{0.123}$\small{$\pm \textbf{0.006}$}} \\
    &        & TabTransformer &  $0.936$ &  $0.218$\small{$\pm 0.006$} &  $0.193$\small{$\pm 0.006$} &  $0.182$\small{$\pm 0.008$} &  $\textbf{0.088}$\small{$\pm \textbf{0.002}$} \\
    &        & VIME &  $0.925$ &  $0.547$\small{$\pm 0.001$} &  $0.582$\small{$\pm 0.001$} &  $0.565$\small{$\pm 0.009$} &  $\textbf{0.492}$\small{$\pm \textbf{0.005}$} \\
\cline{2-8}
    & \multirow{3}{*}{Robust} & RLN &  $0.952$ &  $0.811$\small{$\pm 0.002$} &  $0.821$\small{$\pm 0.001$} &  $0.563$\small{$\pm 0.007$} &  $0.562$\small{$\pm \textbf{0.006}$} \\
    &        & TabTransformer &  $0.939$ &  $0.812$\small{$\pm 0.003$} &  $0.814$\small{$\pm 0.002$} &  $0.568$\small{$\pm 0.008$} &  $\textbf{0.567}$\small{$\pm 0.\textbf{008}$} \\
    &        & VIME &  $0.934$ &  $0.789$\small{$\pm 0.002$} &  $0.796$\small{$\pm 0.002$} &  $0.700$\small{$\pm 0.006$} &  $\textbf{0.697}$\small{$\pm \textbf{0.006}$} \\
\cline{1-8}
\cline{2-8}
\multirow{6}{*}{LCLD} & \multirow{3}{*}{Standard} & RLN &  $0.681$ &  $0.019$\small{$\pm 0.002$} &  $0.265$\small{$\pm 0.007$} &  $0.006$\small{$\pm 0.002$} &  $\textbf{0.001}$\small{$\pm \textbf{0.001}$} \\
    &        & TabTransformer &  $0.706$ &  $0.584$\small{$\pm 0.005$} &  $0.477$\small{$\pm 0.009$} &  $0.150$\small{$\pm 0.007$} &  $\textbf{0.135}$\small{$\pm \textbf{0.006}$} \\
    &        & VIME &  $0.666$ &  $0.137$\small{$\pm 0.005$} &  $0.369$\small{$\pm 0.012$} &  $0.269$\small{$\pm 0.018$} &  $\textbf{0.066}$\small{$\pm \textbf{0.005}$} \\
\cline{2-8}
    & \multirow{3}{*}{Robust} & RLN &  $0.707$ &  $0.692$\small{$\pm 0.001$} &  $0.701$\small{$\pm 0.002$} &  $0.662$\small{$\pm 0.002$} &  $\textbf{0.661}$\small{$\pm \textbf{0.003}$} \\
    &        & TabTransformer &  $0.745$ &  $0.729$\small{$\pm 0.003$} &  $0.735$\small{$\pm 0.002$} &  $0.713$\small{$\pm 0.002$} &  $\textbf{0.708}$\small{$\pm \textbf{0.002}$} \\
    &        & VIME &  $0.654$ &  $0.253$\small{$\pm 0.004$} &  $0.517$\small{$\pm 0.010$} &  $0.226$\small{$\pm 0.008$} &  $\textbf{0.147}$\small{$\pm \textbf{0.005}$} \\
\cline{1-8}
\cline{2-8}
\multirow{6}{*}{CTU} & \multirow{3}{*}{Standard} & RLN &  $0.978$ &  $0.978$\small{$\pm 0.000$} &  $0.978$\small{$\pm 0.000$} &  $\textbf{0.940}$\small{$\pm \textbf{0.002}$} &  $\textbf{0.940}$\small{$\pm \textbf{0.002}$} \\
    &        & TabTransformer &  $\textbf{0.953}$ &  $\textbf{0.953}$\small{$\pm \textbf{0.000}$} &  $\textbf{0.953}$\small{$\pm \textbf{0.000}$} &  $\textbf{0.953}$\small{$\pm \textbf{0.000}$} &  $\textbf{0.953}$\small{$\pm \textbf{0.000}$} \\
    &        & VIME &  $0.951$ &  $0.951$\small{$\pm 0.000$} &  $0.951$\small{$\pm 0.000$} &  $\textbf{0.408}$\small{$\pm \textbf{0.047}$} &  $\textbf{0.408}$\small{$\pm \textbf{0.047}$} \\
\cline{2-8}
    & \multirow{3}{*}{Robust} & RLN &  $0.973$ &  $0.973$\small{$\pm 0.000$} &  $0.973$\small{$\pm 0.000$} &  $\textbf{0.971}$\small{$\pm \textbf{0.000}$} &  $\textbf{0.971}$\small{$\pm \textbf{0.000}$} \\
    &        & TabTransformer &  $\textbf{0.953}$ &  $\textbf{0.953}$\small{$\pm \textbf{0.000}$} &  $\textbf{0.953}$\small{$\pm \textbf{0.000}$} &  $\textbf{0.953}$\small{$\pm \textbf{0.000}$} &  $\textbf{0.953}$\small{$\pm \textbf{0.000}$} \\
    &        & VIME &  $0.951$ &  $0.951$\small{$\pm 0.000$} &  $0.951$\small{$\pm 0.000$} &  $\textbf{0.940}$\small{$\pm \textbf{0.005}$} &  $\textbf{0.940}$\small{$\pm \textbf{0.005}$} \\
\bottomrule
\end{tabular}
\end{table*}

\begin{table*}
\centering
\caption{Attack duration of different attacks in the white-box scenarios A on different models and datasets. Lower is better. The lowest time between MOEVA and CAA is in bold.}
\small
\label{tab:scenario_a_cost}
\begin{tabular}{lll|llll}
\toprule
    &        & Attack &                       CPGD &                       CAPGD &                        MOEVA &                          CAA \\
Dataset & Training & Model &                            &                             &                              &                              \\
\midrule
\multirow{6}{*}{URL} & \multirow{3}{*}{Standard} & RLN &  $1.399$\small{$\pm 2.313$} &   $1.180$\small{$\pm 0.100$} &   $77.277$\small{$\pm 2.909$} &   $\textbf{16.463}$\small{$\pm \textbf{0.850}$} \\
    &        & TabTransformer &  $1.723$\small{$\pm 2.722$} &   $1.491$\small{$\pm 0.346$} &   $77.488$\small{$\pm 2.736$} &   $\textbf{13.695}$\small{$\pm \textbf{0.582}$} \\
    &        & VIME &  $0.359$\small{$\pm 0.309$} &   $1.091$\small{$\pm 0.158$} &   $76.024$\small{$\pm 5.359$} &   $\textbf{44.449}$\small{$\pm \textbf{1.878}$} \\
\cline{2-7}
    & \multirow{3}{*}{Robust} & RLN &  $1.483$\small{$\pm 2.393$} &   $1.147$\small{$\pm 0.213$} &   $78.514$\small{$\pm 2.445$} &   $\textbf{66.551}$\small{$\pm \textbf{1.771}$} \\
    &        & TabTransformer &  $0.528$\small{$\pm 0.197$} &   $1.152$\small{$\pm 0.312$} &   $72.602$\small{$\pm 2.785$} &   $\textbf{62.486}$\small{$\pm \textbf{2.337}$} \\
    &        & VIME &  $1.247$\small{$\pm 2.003$} &   $2.803$\small{$\pm 3.520$} &   $74.944$\small{$\pm 3.478$} &   $\textbf{61.597}$\small{$\pm \textbf{2.258}$} \\
\cline{1-7}
\cline{2-7}
\multirow{6}{*}{LCLD} & \multirow{3}{*}{Standard} & RLN &  $2.454$\small{$\pm 1.533$} &   $1.368$\small{$\pm 0.053$} &   $54.033$\small{$\pm 2.345$} &    $\textbf{5.530}$\small{$\pm \textbf{0.360}$} \\
    &        & TabTransformer &  $1.309$\small{$\pm 1.328$} &   $1.387$\small{$\pm 0.193$} &  $126.230$\small{$\pm 6.921$} &   $\textbf{73.334}$\small{$\pm \textbf{4.383}$} \\
    &        & VIME &  $2.602$\small{$\pm 1.915$} &   $1.405$\small{$\pm 0.547$} &   $51.689$\small{$\pm 2.906$} &   $\textbf{10.906}$\small{$\pm \textbf{0.933}$} \\
\cline{2-7}
    & \multirow{3}{*}{Robust} & RLN &  $2.339$\small{$\pm 1.787$} &   $1.579$\small{$\pm 0.172$} &   $59.459$\small{$\pm 3.655$} &   $\textbf{51.677}$\small{$\pm \textbf{2.255}$} \\
    &        & TabTransformer &  $1.107$\small{$\pm 0.917$} &   $1.454$\small{$\pm 0.110$} &  $129.317$\small{$\pm 7.567$} &  $\textbf{125.970}$\small{$\pm \textbf{5.223}$} \\
    &        & VIME &  $1.423$\small{$\pm 0.793$} &   $1.364$\small{$\pm 0.263$} &   $52.615$\small{$\pm 2.413$} &   $\textbf{19.454}$\small{$\pm \textbf{1.192}$} \\
\cline{1-7}
\cline{2-7}
\multirow{6}{*}{CTU} & \multirow{3}{*}{Standard} & RLN &  $1.313$\small{$\pm 0.080$} &  $11.230$\small{$\pm 0.291$} &  $\textbf{102.815}$\small{$\pm \textbf{2.219}$} &  $114.721$\small{$\pm 2.640$} \\
    &        & TabTransformer &  $1.385$\small{$\pm 0.099$} &  $11.029$\small{$\pm 0.157$} &  $\textbf{105.283}$\small{$\pm \textbf{2.478}$} &  $115.654$\small{$\pm 3.364$} \\
    &        & VIME &  $1.340$\small{$\pm 0.016$} &  $10.890$\small{$\pm 0.148$} &  $\textbf{111.760}$\small{$\pm \textbf{3.860}$} &  $116.981$\small{$\pm 2.015$} \\
\cline{2-7}
    & \multirow{3}{*}{Robust} & RLN &  $1.362$\small{$\pm 0.062$} &  $15.850$\small{$\pm 9.147$} &  $\textbf{101.560}$\small{$\pm \textbf{2.005}$} &  $114.210$\small{$\pm 2.378$} \\
    &        & TabTransformer &  $1.310$\small{$\pm 0.011$} &  $10.892$\small{$\pm 0.253$} &  $\textbf{104.715}$\small{$\pm \textbf{2.519}$} &  $115.586$\small{$\pm 1.449$} \\
    &        & VIME &  $1.376$\small{$\pm 0.034$} &  $11.063$\small{$\pm 0.165$} &  $\textbf{112.187}$\small{$\pm \textbf{1.123}$} &  $120.314$\small{$\pm 3.053$} \\
\bottomrule
\end{tabular}
\end{table*}

We compare the robust accuracy after each of the constrained attacks CPGD, CAPGD, MOEVA, and CAA for the standard and robust models across all our datasets in Table \ref{tab:scenario_a_effectiveness}. Against non-robust models (dashed lines), CAA outperforms all the attacks in 6 over 9 cases. It is equally performing as the best attack (MOEVA) in the three remaining cases of CTU. 

Against adversarially trained models (continuous lines), CAA and MOEVA achieve similar performances across 8/9 combinations. CAA outperforms MOEVA for VIME architecture trained on LCLD.

The main advantage of CAA is its ability to find "easy" constrained adversarial using cheaper gradient attacks. We compare in Table \ref{tab:scenario_a_cost} the cost of running each of the four attacks. CAA particularly shines in terms of efficiency. 
It is significantly less expensive than MOEVA alone for LCLD and URL datasets, and marginally more costly for the CTU dataset (the most robust model). For example, the cost of attacking standard URL models is divided by three for TabTransformer and RLN models and divided by five for RLN and VIME standard models on the LCLD dataset.

\subsection{Impact of Defender's Capabilities}

\begin{table}[]
    \centering
    {\small
    \caption{Robustness to CAA adversarial attacks. S: Strong, M: Medium, W: Weak.}
    \label{tab:comapre_architecture}
    \begin{tabular}{l|ccc|ccc}
    \toprule
        Training &  \multicolumn{3}{c}{Standard} & \multicolumn{3}{c}{Adversarial} \\
        Dataset         & URL   & LCLD  & CTU   & URL   & LCLD  & CTU \\
        \hline
        TabTr  &  W     & W      & S      &  M     &  S     & S\\
        RLN             &  W     & W      &  S     &  M     &  S     & S \\
        VIME            &  W     &  W     &  W     &  M     &   W    & S\\
         \bottomrule
    \end{tabular}
    }
    
\end{table}

\paragraph{Test performances can be misleading when selecting an architecture.}
The three architectures we evaluate all achieve similar test performances across our three datasets (Table \ref{tab:data}), and previous surveys \cite{Borisov_2022} have shown their equivalence across multiple tasks. Our results show however that their robustness to adversarial noise varies significantly across our use cases.

We summarize in Table \ref{tab:comapre_architecture} the robust accuracy of each architecture against our white-box adversarial attacks.
Without adversarial training, TabTransformer and RLN are completely robust on the CTU dataset, and completely vulnerable against URL and LCLD. With adversarial training, they become robust on LCLD but remain vulnerable on the URL dataset.
VIME is vulnerable to adversarial attacks across our three datasets with standard training and remains vulnerable on LCLD even with adversarial training. 

\paragraph{Adversarial training with unconstrained examples is sufficient.}

Adversarial training consistently improves the robustness of our models across all the datasets and architectures even if the adversarial examples generated are not constrained. 
Unconstrained adversarial examples are cheaper than constrained examples, which are even cheaper than problem space examples as demonstrated by \cite{dyrmishi2022empirical}. Our benchmark confirms the cheapest adversarial examples are sufficient for tabular machine learning against white-box attacks.

\paragraph{The most robust architecture against transfer attacks is TabTransformer.} 
We compare in Table \ref{tab:surrogate_ranking_main} the ranking of each architecture in terms of robust accuracy when used as a surrogate source model (by the attacker), and when used as target (the defender) for each of our scenarios C, D, and E.
The lower the rank the better for the attacker, and the higher the rank the better for the defender.
Our results show that for robust targets TabTransformer is the best architecture for the defender, and VIME the worst. There is however no architecture that is reliably best suited to use as a surrogate source against standard and robust targets.


\begin{table}
\small
    \centering
    \caption{Rank of target robust accuracy when using different architectures as source with standard training (the lowest rank best for attacker), and as target (the highest rank the best for defender). All models are trained for the LCLD dataset.}
    \begin{tabular}{c|c|lll|lll}
    \toprule
              &                & \multicolumn{3}{c|}{As a source $\downarrow$}                                          & \multicolumn{3}{c}{As a target $\uparrow$  }                                          \\ 
 & Scenario       & \multicolumn{1}{c}{C1} & \multicolumn{1}{c}{D1} & \multicolumn{1}{c|}{E1} & \multicolumn{1}{c}{C1} & \multicolumn{1}{c}{D1} & \multicolumn{1}{c}{E1} \\
Target              &       Model  & & & & & & \\ \midrule
Standard            & TabTransf. & 1.5                    & 2                      & 2                       & 1                      & 2                      & 1.5                    \\
                & RLN            & 1.5                    & 1                      & 1                       & 1.5                    & 1                      & 1                      \\
                & VIME           & 1.5                    & 1.5                    & 1.5                     & 2                      & 1.5                    & 2                      \\ \midrule
Robust                & TabTransf. & 1.5                    & 2                      & 1.5                     & 2                      & 2                      & 2                      \\
                & RLN            & 1.5                    & 1.5                    & 1                       & 1.5                    & 1.5                    & 1.5                    \\
                & VIME           & 1.5                    & 1                      & 2                       & 1                      & 1                      & 1 \\ \bottomrule                    
\end{tabular}
    
    \label{tab:surrogate_ranking_main}
\end{table}

\subsection{Impact of Attacker's Capabilities}

In this section, we evaluate the impact of each of the three capabilities that influence the success rate of the attack as described in Section \ref{sec:threat_models}: access to the target model, access to the training dataset, and access to the domain knowledge.

\paragraph{Some architectures remain robust even with increased access to the target model.} 
We evaluate the impact of increasing the query budgets of MOEVA (scenario B1) on the robustness of deep tabular models.
In this scenario, the number of calls to the model is the product of the number of search generations and the number of offsprings at each generation. 

We evaluate the robust accuracy of the models trained on the LCLD, URL, and CTU datasets in Table \ref{tab:scenario_b_cost}. We study both the case where models have standard training (a) and the case where they are adversarially trained (b).

Across all our LCLD models, increasing the budget of the attack does not significantly impact its effectiveness. Robust models (adversarially trained RLN and TabTransformer) remain robust under a four times budget increase.

The robustness of models trained on the URL dataset on the contrary is very sensitive to the attack budgets. For robust RLN and TabTransformer, the accuracy is halved when the budget is quadrupled (from 100x100 to 200x200).

\begin{table*}[]
\centering
\caption{Model robustness in the scenarios B1 for our three datasets. The budget is (\#generation)x(\#offspring per generation). A lower robust accuracy means a more effective attack. The lowest robust accuracy is in bold.}
\label{tab:scenario_b_cost}
\footnotesize

\begin{tabular}{@{}ll|lllllllll@{}}
\toprule
Dataset &       & \multicolumn{3}{c}{URL}                                                              & \multicolumn{3}{c}{LCLD}                                                             & \multicolumn{3}{c}{CTU}                                                              \\
Model        &         & RLN                        & TabTrans.             & VIME                       & RLN                        & TabTrans.             & VIME                       & RLN                        & TabTrans.             & VIME                       \\
Training            & Budget  &                            &                            &                            &                            &                            &                            &                            &                            &                            \\ \midrule
\multirow{5}{*}{Standard} & 50x50   & $0.612$ & $0.577$ & $0.785$ & $0.017$ & $0.258$ & $0.281$ & $0.964$ & $\textbf{0.953}$ & $0.888$ \\
                          & 100x100 & $0.236$ & $0.182$ & $0.565$ & $\textbf{0.006}$ & $\textbf{0.150}$ & $\textbf{0.269}$ & $0.940$ & $\textbf{0.953}$ & $0.408$ \\
                          & 100x200 & $0.162$ & $0.093$ & $0.368$ & $0.042$ & $0.160$ & $0.308$ & $0.303$ & $\textbf{0.953}$ & $0.088$ \\
                          & 200x100 & $0.154$ & $0.087$ & $0.353$ & $0.053$ & $0.169$ & $0.315$ & $0.241$ & $\textbf{0.953}$ & $0.060$ \\
                          & 200x200 & $\textbf{0.134}$ & $\textbf{0.056}$ & $\textbf{0.293}$ & $0.076$ & $0.179$ & $0.332$ & $\textbf{0.010}$ & $\textbf{0.953}$ & $\textbf{0.013}$ \\ \midrule
\multirow{5}{*}{Robust}   & 50x50   & $0.834$ & $0.827$ & $0.836$ & $0.673$ & $0.717$ & $0.314$ & $0.972$ & $\textbf{0.953}$ & $0.949$ \\
                          & 100x100 & $0.563$ & $0.568$ & $0.700$ & $\textbf{0.662}$ & $\textbf{0.713}$ & $\textbf{0.226}$ & $0.971$ & $\textbf{0.953}$ & $0.940$ \\
                          & 100x200 & $0.431$ & $0.445$ & $0.620$ & $0.663$ & $0.715$ & $0.240$ & $0.970$ & $\textbf{0.953}$ & $0.928$ \\
                          & 200x100 & $0.424$ & $0.441$ & $0.613$ & $0.664$ & $0.715$ & $0.247$ & $0.969$ & $\textbf{0.953}$ & $0.924$ \\
                          & 200x200 & $\textbf{0.403}$ & $\textbf{0.415}$ & $\textbf{0.589}$ & $0.664$ & $0.714$ & $0.249$ & $\textbf{0.968}$ & $\textbf{0.953}$ & $\textbf{0.907}$ \\ \bottomrule
\end{tabular}
\end{table*}

\begin{figure*}
    \begin{subfigure}[b]{0.49\textwidth}
    \centering
    \includegraphics[width=\textwidth]{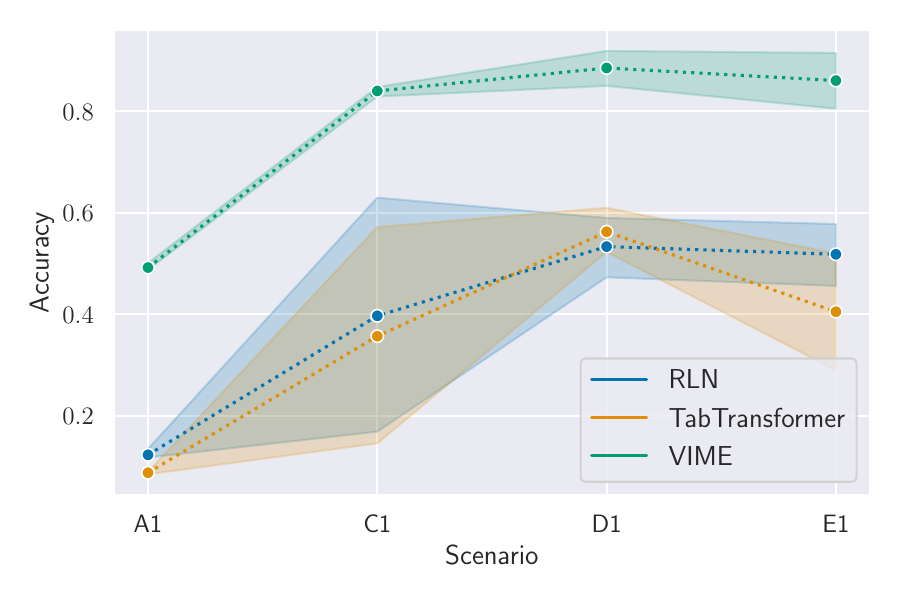}
    \caption{Standard URL}
    \end{subfigure}
    \begin{subfigure}[b]{0.49\textwidth}
    \centering
    \includegraphics[width=\textwidth]{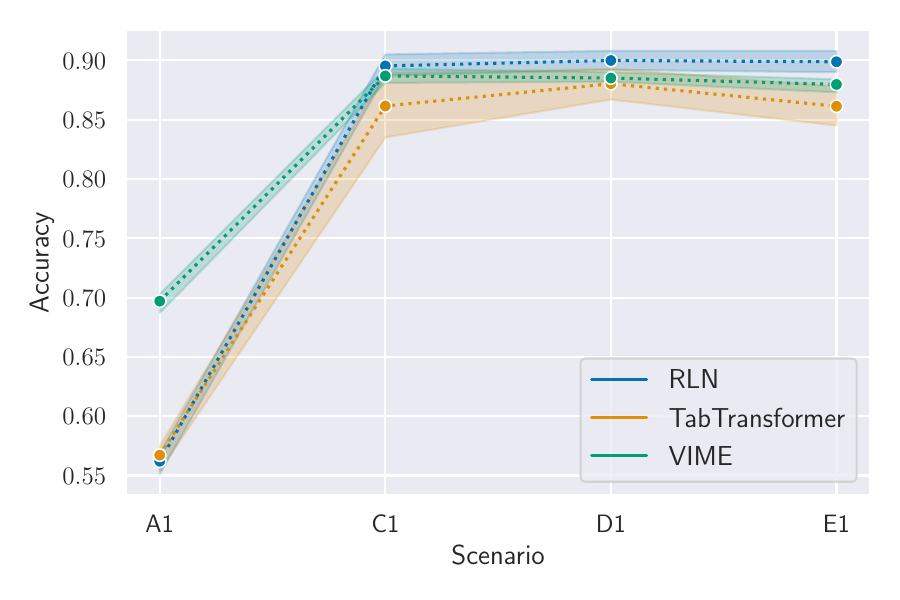}
    \caption{Robust URL}
    \end{subfigure}

    \begin{subfigure}[b]{0.49\textwidth}
    \centering
    \includegraphics[width=\textwidth]{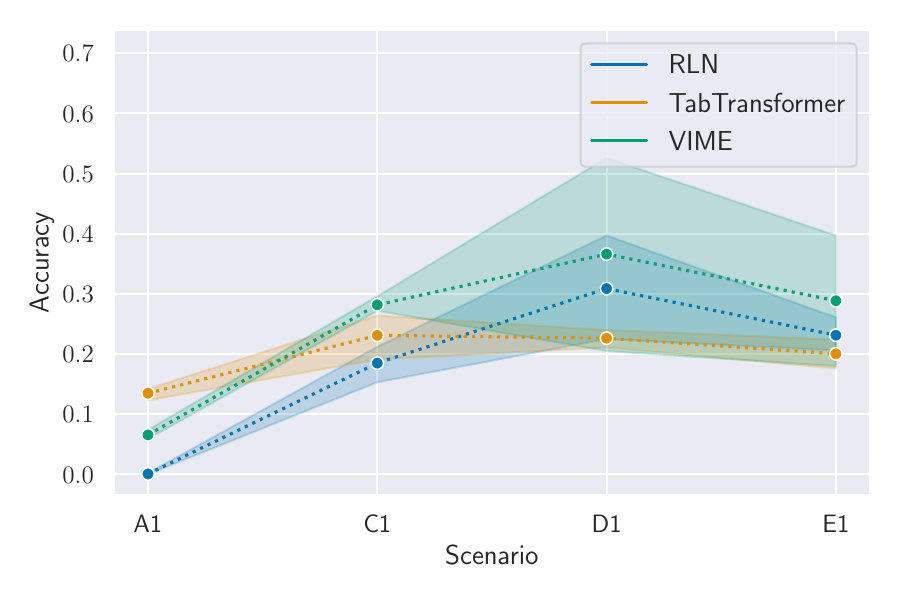}
    \caption{Standard LCLD}
    \end{subfigure}
    \begin{subfigure}[b]{0.49\textwidth}
    \centering
    \includegraphics[width=\textwidth]{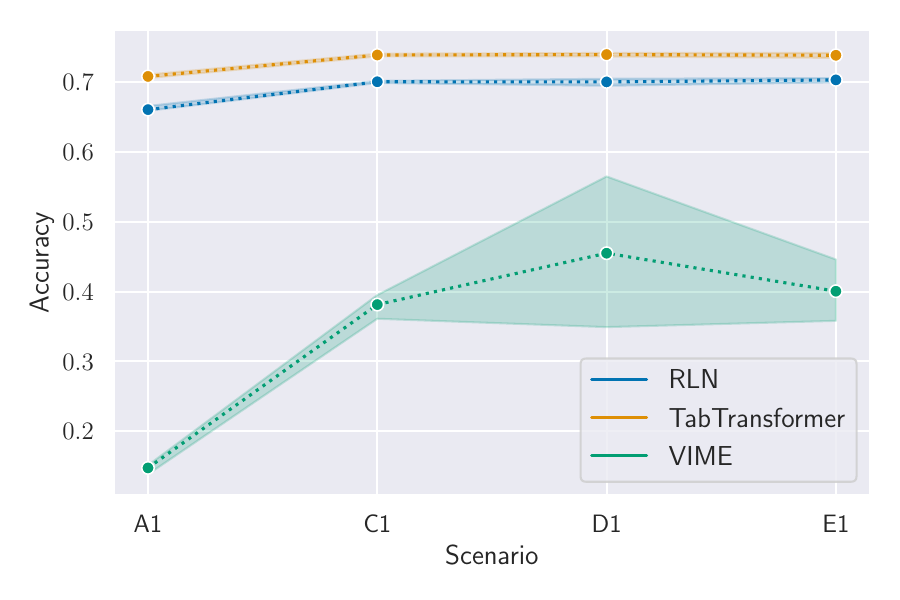}
    \caption{Robust LCLD}
    \end{subfigure}

    \begin{subfigure}[b]{0.49\textwidth}
    \centering
    \includegraphics[width=\textwidth]{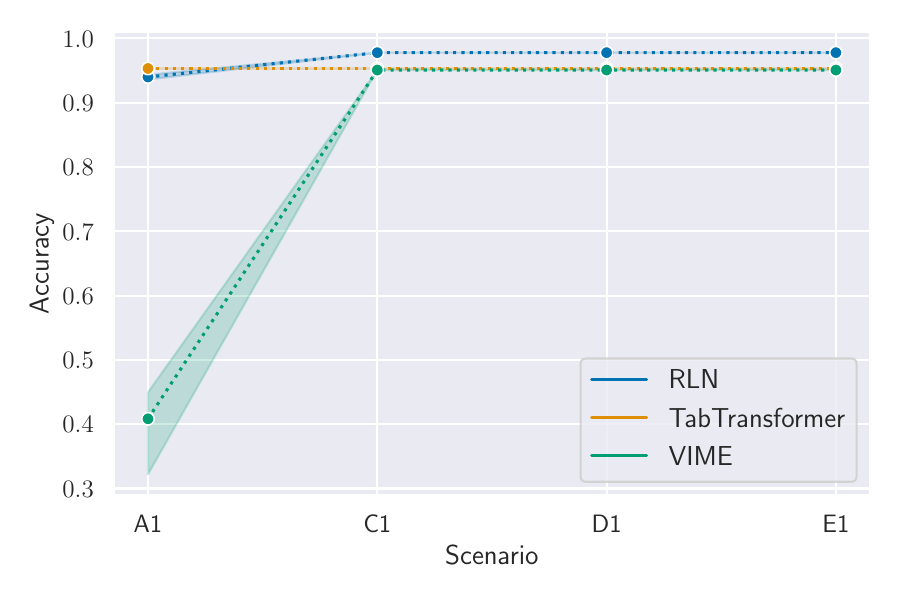}
    \caption{Standard CTU}
    \end{subfigure}
    \begin{subfigure}[b]{0.49\textwidth}
    \centering
    \includegraphics[width=\textwidth]{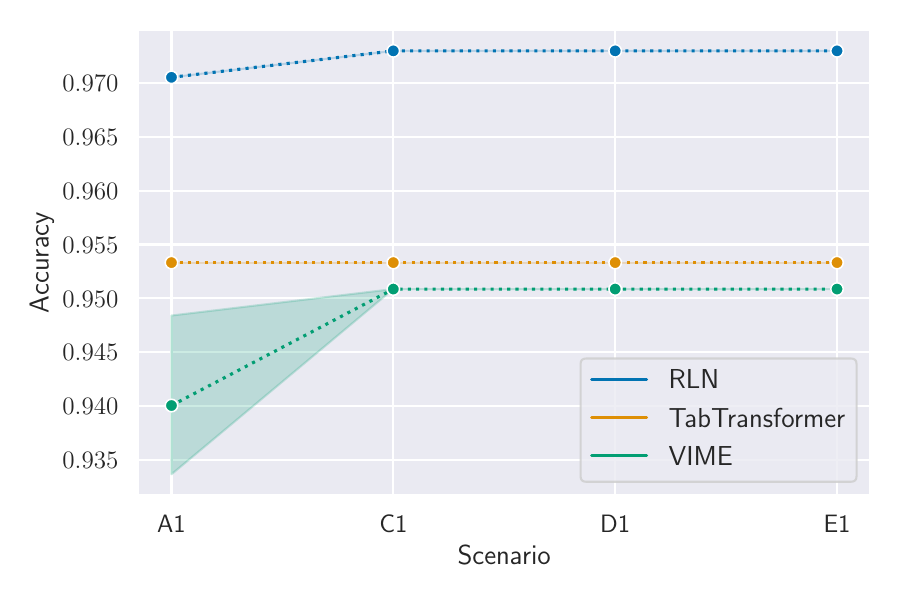}
    \caption{Robust CTU}
    \end{subfigure}

    \caption{Robust accuracy for different scenarios and target architectures trained with standard training (left) and adversarial training (right). The range represents the min-max interval, over 5 seeds for A, and over 5 seeds and 2 source models (different from the target model) for C D E.}
    \label{fig:scenario_cde}
\end{figure*}

\begin{figure*}
    \begin{subfigure}[b]{0.49\textwidth}
    \centering
    \includegraphics[width=\textwidth]{figures/lcld_v2_iid/acde_Standard_1.pdf}
    \caption{Standard target (A1, C1, D1, E1)}
    \end{subfigure}
    \begin{subfigure}[b]{0.49\textwidth}
    \centering
    \includegraphics[width=\textwidth]{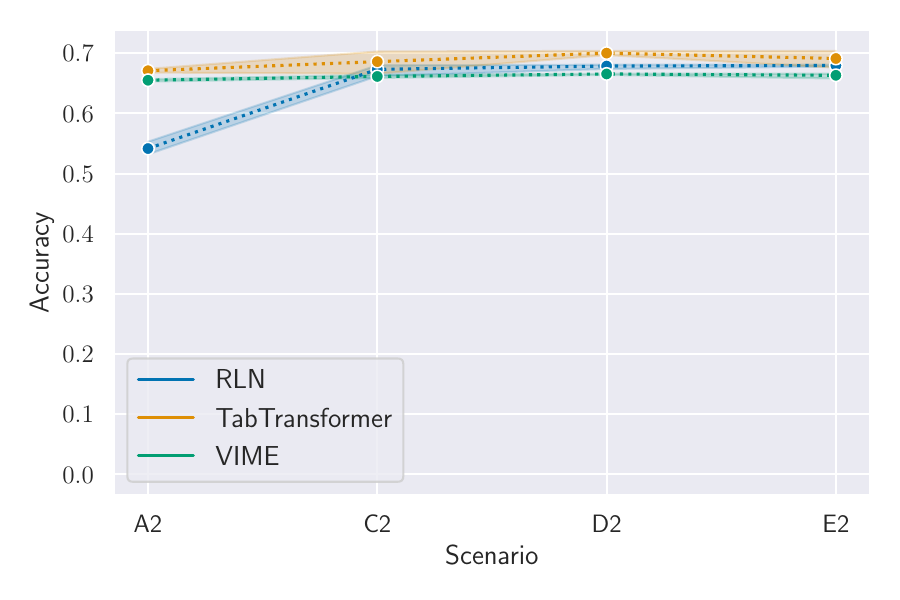}
    \caption{Standard target (A2, C2, D2, E2)}
    \end{subfigure}
    \begin{subfigure}[b]{0.49\textwidth}
    \centering
    \includegraphics[width=\textwidth]{figures/lcld_v2_iid/acde_Robust_1.pdf}
    \caption{Robust target (A1, C1, D1, E1)}
    \end{subfigure}
    \begin{subfigure}[b]{0.49\textwidth}
    \centering
    \includegraphics[width=\textwidth]{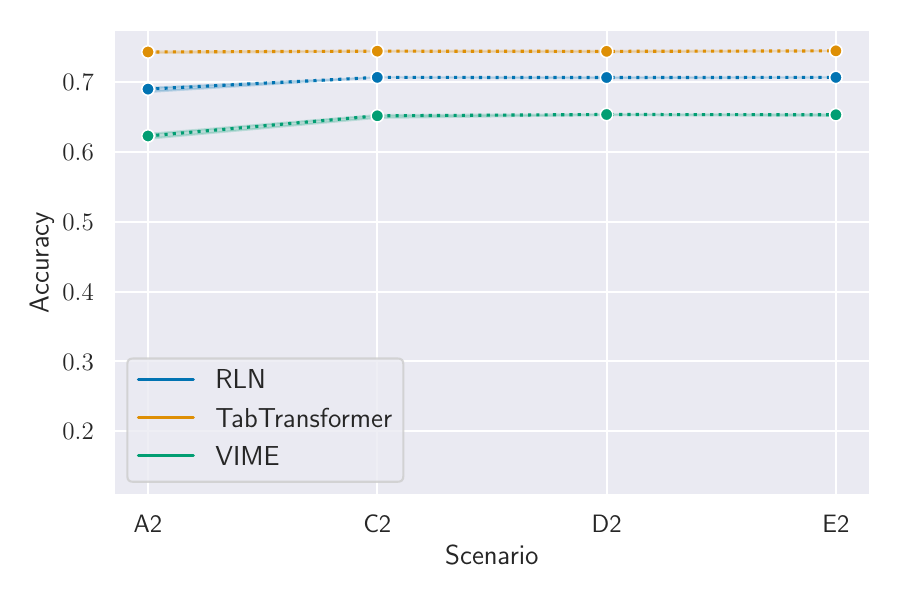}
    \caption{Robust target (A2, C2, D2, E2)}
    \end{subfigure}

    \caption{Robust accuracy for different scenarios and target models. C1/C2 source is not robust. The range represents the min-max interval, over 5 seeds for A, and over 5 seeds and 2 source models (different from the target model) for C D E.}
    \label{fig:scenario_cde_lcld}
\end{figure*}

\paragraph{The attacker does not need access to the training data for successful attacks.} 
We evaluate respectively in scenarios C, D, and E transfer attacks where the attacker has full access to the training set, 10\% of the training set, and a set of the same distribution as the training. 
We study the different scenarios for our three datasets in Figure \ref{fig:scenario_cde}.

For CTU (bottom), while some architectures can be vulnerable to white-box scenarios (scenario A), all the architectures are robust to transfer scenarios, even when the full training set is collected by the attacker (scenario C). 

For URL and LCLD, our study shows that models trained with standard training are vulnerable to all transfer attacks, even under scenario E where the attacker doesn't have access to any training point.
Adversarial training fixes this vulnerability except for VIME on the LCLD dataset. This combination remains vulnerable against all transfer attacks.

\paragraph{The attacker absolutely needs domain knowledge for successful attacks.} 
For each of our scenarios, we evaluated two variants. In scenarios A1, C1, D1, and E1, the attacker has access to some domain knowledge of the problem at hand. The attacker is aware of the constraints between the features, their boundaries, and their immutability.
In scenarios A2, C2, D2, and E2, the attacker does not have such knowledge. 

We evaluate in Figure \ref{fig:scenario_cde_lcld} the impact of domain knowledge on the success rate of the different attacks. We cover in this Figure the LCLD dataset, but the same conclusions apply across all our datasets.
The attacker in Figures \ref{fig:scenario_cde_lcld} (b) and (d) is not aware of the domain constraints and only maximizes the classification error. Our results show that the models are completely robust in transfer attacks without this crucial information but with standard and adversarial training.

\section{Perspectives}

Our work was designed to uplift the research on adversarial robustness for tabular deep learning models. It raises numerous opportunities to explore further realistic threat scenarios and improve the models currently deployed for critical applications. In particular, we propose four lines of investigation that we believe would have the most impact on practitioners in critical machine learning.  

\paragraph{Beyond binary classification.} All our models tackle binary classification, but the algorithms we designed and the multi-scenario evaluation protocol we presented remain relevant for multi-class classification. Our evaluation focused on binary classification because this task matches real-world scenarios of ML systems in production. For financial ML, credit scoring and fraud detection are among the most popular applications of ML and are binary classification tasks. In computer security, phishing, malware, and botnet detection are binary classification tasks. Multi-class applications open new threat models by introducing concepts like targeted attacks that can be relevant in a few critical settings.

\paragraph{Non $L_2$ norms perturbations.} Tabular Machine Learning involves categorical features for which a $L_2$ distance norm is the most relevant distance. While there are ongoing studies that envision other notions of imperceptibility~\cite{wang2022similaritybased,debenedetti2023evading}, $L_2$ distance has been studied in the work by Dyrmishi et al. \cite{dyrmishi2022empirical} and their evaluation supported the relevance of this setting for our use cases.

\paragraph{Higher order constraints.} 
Our repair operators only handle equality relationships with a single feature on the left-hand side of the definition.
Repairing inequality constraints and multi-features equality constraints equation remains an open problem. However, empirically, our algorithms already manage to satisfy these constraints for most of the inputs. In Table \ref{tab:ctr_satisfaction}, we compare the constraint satisfactions of the adversarial examples generated with each algorithm. Each row is one of the nine constraints of the LCLD dataset. The nine constraints are mostly satisfied across the different attacks. 
Provable satisfaction of higher-order constraints is an ambitious endeavor, for which our study provides both relevant datasets and evaluation protocols.   

\paragraph{More models and datasets.} Our benchmark is a live benchmark with new models and datasets that are updated regularly and available at the artifact URL\footnote{https://figshare.com/s/10d0af55d00e7f27f142}. 
We are also opening a public leaderboard where users can propose new constrained datasets and submit their own models to be evaluated across the different threat models of our protocol\footnote{Public leaderboard disclosed after review to ensure double blind}.

\begin{table}
\centering
\caption{Constraint Satisfaction for LCLD dataset classification on TabTransformer.}
\small
\begin{tabular}{ll|lll}
\toprule
           & Attack &             CPGD & CAPGD & MOEVA \\
Constraint & Constraints &                  &       &       \\
\midrule

$\phi_{1}$ & $\land, \leq$ &            0.960 & 0.953 & 0.975 \\
$\phi_{2}$ & $\leq$ &            0.990 & 1.000 & 1.000 \\
$\phi_{3}$ & $\leq$ &            1.000 & 1.000 & 1.000 \\
$\phi_{4}$ & $\in$ &            1.000 & 1.000 & 1.000 \\
$\phi_{5}$ & $=$ &            1.000 & 1.000 & 0.317 \\
$\phi_{6}$ & $=$ &            1.000 & 1.000 & 0.082 \\
$\phi_{7}$ & $=$ &            1.000 & 1.000 & 1.000 \\
$\phi_{8}$ & $=$ &            1.000 & 1.000 & 1.000 \\
$\phi_{9}$ & $=$ &            1.000 & 1.000 & 1.000 \\
\bottomrule
\end{tabular}
\label{tab:ctr_satisfaction}
\end{table}

\section*{Conclusion}

In this work, we propose a realistic evaluation of the robustness of deep tabular machine learning models in ten scenarios with different attacker capabilities. Our scenarios consider an attacker with different levels of knowledge of the model architecture, the training data, and the domain of application.
We propose a new adaptive attack that can handle the specificities of tabular machine learning, in particular domain constraints. It combines the efficiency of gradient-based attacks and  the effectiveness of a search-based methods. Our new attack is as effective as the best-performing competing attack while being 5 times more efficient.
Our contribution enables a fine-grained evaluation of state-of-the-art deep learning models and reveals the importance of knowledge about the target distribution and domain constraints. 
From a defender's perspective, we showed that limiting testing to clean performance can be misleading, as the most effective architecture on clean data is not necessarily the most robust. 
Additionally, we showed that adversarial training with unconstrained examples is sufficient to defend against constrained adversarial examples. 
From an attacker perspective, we showed that knowledge of domain constraints, in particular the relationship between features is essential to generate valid adversarial examples.
On the other hand, we showed that an attacker does not need the model architecture nor the training data but can use a surrogate and a training set of the same distribution to generate adversarial examples.

%

\bibliographystyle{plain}
\bibliography{adv,tabsurvey,blackbox}

\clearpage
\section*{Appendix \& Availability}
\appendix
\section{Experimental protocol}

\subsection{Datasets}

\begin{table*}[t]
  \centering
  \caption{The datasets evaluated in the empirical study, with the class imbalance of each dataset.}
  \label{tab:data_extended}
  \small
  \begin{tabular}{c|llll|lll}
    \toprule
    Dataset & \multicolumn{4}{c}{Properties} & \multicolumn{3}{|c}{Test performances (AUC)}\\
     & Task & Size & \# Features & Balance (\%) & TabTransformer & RLN & VIME\\
    \midrule
    LCLD~\cite{lcld} &  Credit Scoring &  1 220 092 & 28 & 80/20 & 0.717 & 0.719 & 0.714 \\
    CTU-13~\cite{chernikova2022fence}&     Botnet Detection    &   198 128 & 756 & 99.3/0.7 & 0.981 & 0.984 & 0.974 \\
    URL~\cite{hannousse2021towards} & Phishing URL detection    & 11 430 & 63  & 50/50 & 0.979 & 0.991 & 0.987 \\
    \bottomrule
  \end{tabular}
\end{table*}

Our dataset design followed the same protocol as Simonetto et al.\cite{simonetto2021unified}.
We present in Table \ref{tab:data_extended} the attributes of our datasets and the test performance achieved by each of the architectures.

\paragraph{Credit scoring - LCLD:}
We engineer a dataset from the publicly available Lending Club Loan Data (\textit{https://www.kaggle.com/wordsforthewise/lending-club}).
This dataset contains 151 features, and each example represents a loan that was accepted by the Lending Club.
However, among these accepted loans, some are not repaid and charged off instead.
Our goal is to predict, at the request time, whether the borrower will be repaid or charged off.
This dataset has been studied by multiple practitioners on Kaggle. 
However, the original version of the dataset contains only raw data and to the extent of our knowledge, there is no featured engineered version commonly used.
In particular, one shall be careful when reusing feature-engineered versions, as most of the versions proposed presents data leakage in the training set that makes the prediction trivial.
Therefore, we propose our own feature engineering. 
The original dataset contains 151 features. 
We remove the example for which the feature ``loan status'' is different from ``Fully paid'' or  ``Charged Off'' as these represent the only final status of a loan: for other values, the outcome is still uncertain. For our binary classifier, a `Fully paid'' loan is represented as 0 and a ``Charged Off'' as 1.
We start by removing all features that are not set for more than 30\% of the examples in the training set. 
We also remove all features that are not available at loan request time, as this would introduce bias. 
We impute the features that are redundant (e.g. grade and sub-grade) or too granular (e.g. address) to be useful for classification.
Finally, we use one-hot encoding for categorical features.
We obtain 47 input features and one target feature.
We split the dataset using random sampling stratified on the target class and obtain a training set of 915K examples and a testing set of 305K.
They are both unbalanced, with only 20\% of charged-off loans (class 1). 
We trained a neural network to classify accepted and rejected loans. It has 3 fully connected hidden layers with 64, 32, and 16 neurons.

For each feature of this dataset, we define boundary constraints as the extremum value observed in the training set.
We consider the 19 features that are under the control of the Lending Club as immutable. 
We identify 10 relationship constraints (3 linear, and 7 non-linear ones).

\paragraph{Botnet attacks - CTU-13:}
This is a feature-engineered version of CTU-13 proposed by~\cite{chernikova2019fence}.
It includes a mix of legit and botnet traffic flows from the CTU University campus. Chernikova et al. aggregated the raw network data related to packets, duration, and bytes for each port from a list of commonly used ports. 
The dataset is made of 143K training examples and 55K testing examples, with 0.74\% examples labeled in the botnet class (traffic that a botnet generates). Data have 756 features, including 432 mutable features. We identified two types of constraints that determine what feasible traffic data is. The first type concerns the number of connections and requires that an attacker cannot decrease it. The second type is inherent constraints in network communications (e.g. maximum packet size for TCP/UDP ports is 1500 bytes). In total, we identified 360 constraints.

\paragraph{URL Phishing - ISCX-URL2016:}

Phishing attacks are usually used to conduct cyber fraud or identity theft.
This kind of attack takes the form of a URL that reassembles a legitimate URL (e.g. user's favorite e-commerce platform) but redirects to a fraudulent website that asks the user for their personal or banking data. 
\cite{hannousse2021towards} extracted features from legitimate and fraudulent URLs as well as external service-based features to build a classifier that can differentiate fraudulent URLs from legitimate ones.
The feature extracted from the URL includes the number of special substrings such as ``www", ``\&", ``,", ``\$", "and", the length of the URL, the port, the appearance of a brand in the domain, in a subdomain or in the path, and the inclusion of ``http" or ``https".
External service-based features include the Google index, the page rank, and the presence of the domain in the DNS records.
The complete list of features is present in the reproduction package.
\cite{hannousse2021towards} provide a dataset of 5715 legit and 5715 malicious URLs.
We use 75\% of the dataset for training and validation and the remaining 25\% for testing and adversarial generation. 

We extract a set of 14 relation constraints between the URL features.
Among them, 7 are linear constraints (e.g. length of the hostname is less or equal to the length of the URL) and 7 are Boolean constraints of the type $if a > 0 $ then $ b > 0$ (e.g. if the number of http $>$ 0 then the number slash ``/" $>$ 0).

\subsection{Model architectures}
\begin{table}
\centering
  \caption{The three model architectures of our study.}
  \label{tab:models}
  \small
  \begin{tabular}{lll}
  
    \toprule
    Family & Model & Hyperparameters\\
    \midrule
    Transformer & TabTransformer  & \begin{tabular}[c]{@{}l@{}}$hidden\_dim$, $n\_layers$\\ $learning\_rate$, $norm$, $\theta$\end{tabular}  \\
    Regularization & RLN & \begin{tabular}[c]{@{}l@{}}$hidden\_dim$, $depth$, \\ $heads$, $weight\_decay$, \\ $learning\_rate$, $dropout$\end{tabular} \\
        Encoding & VIME  &  $p_m$, $\alpha$, $K$, $\beta$\\

  \bottomrule
\end{tabular}
\end{table}

Table \ref{tab:models} summarizes the family, model architecture, and hyperparameters tuned during training of our models.

\subsection{Implementation and hardware}
For the C-PGD approach, we extend the implementation of PGD proposed by TorchAttacks\cite{Kim2020Torchattacks:Attacks}.
We also extend the attack to build our CAPGD and CAA attacks.

We run our experiments on an HPC cluster node with 32 cores and 64GB of RAM dedicated to our task.
Each node consists of 2 AMD Epyc ROME 7H12 @ 2.6 GHz for a total of 128 cores with 256 GB of RAM.
\section{Detailed results}

In the following, we provide the detailed results of the figures in the main paper.

\subsection{Which defender’s capabilities impact the success rate?}

\paragraph{The most robust architecture against transfer attacks is TabTransformer.}

We compared in Table \ref{tab:surrogate_ranking_main} the ranking of each architecture in terms of robust accuracy when used as a surrogate source model (by the attacker), and when used as target (the defender) for each of our scenarios C, D, and E in the case of LCLD dataset.

In the following, we present additional results for URL and CTU datasets.
For both URL and CTU, the best architecture for robust defender is CTU, and the weakest is VIME for URL.

\begin{table}[h]
\small
    \centering
 \begin{tabular}{c|c|lll|lll}
                \toprule
              &                & \multicolumn{3}{c|}{As a source $\downarrow$}                                          & \multicolumn{3}{c}{As a target $\uparrow$  }                                          \\ 
 & Scenario       & \multicolumn{1}{c}{C1} & \multicolumn{1}{c}{D1} & \multicolumn{1}{c|}{E1} & \multicolumn{1}{c}{C1} & \multicolumn{1}{c}{D1} & \multicolumn{1}{c}{E1} \\
Target              &       Model  & & & & & & \\ \midrule
Standard                & TabTransf. & 1                      & 1                      & 1                       & 1                      & 1                      & 1                      \\
                & RLN            & 1.5                    & 2                      & 1.5                     & 1.5                    & 1.5                    & 1.5                    \\
                & VIME           & 2                      & 1.5                    & 2                       & 2                      & 2                      & 2                      \\ \midrule
Robust                & TabTransf. & 1.5                    & 1                      & 1.5                     & 1                      & 1                      & 1                      \\
                & RLN            & 1                      & 1.5                    & 1.5                     & 1.5                    & 2                      & 2                      \\
                & VIME           & 2                      & 2                      & 1.5                     & 2                      & 1.5                    & 1.5                   \\ \bottomrule
\end{tabular}
    \caption{URL - Rank of target robust accuracy when using different architectures as source with standard training (the lowest rank best for attacker), and as target (the highest rank the best for defender) }
    \label{tab:surrogate_ranking_url}
\end{table}

\begin{table}[h]
\small
\centering
 \begin{tabular}{c|c|lll|lll}
    \toprule
              &                & \multicolumn{3}{c|}{As a source $\downarrow$}                                          & \multicolumn{3}{c}{As a target $\uparrow$  }                                          \\ 
 & Scenario       & \multicolumn{1}{c}{C1} & \multicolumn{1}{c}{D1} & \multicolumn{1}{c|}{E1} & \multicolumn{1}{c}{C1} & \multicolumn{1}{c}{D1} & \multicolumn{1}{c}{E1} \\
Target              &       Model  & & & & & & \\ \midrule
Standard                & TabTransf. & 1.5                    & 1.5                    & 1.5                     & 1.5                    & 1.5                    & 1.5                    \\
                & RLN            & 1.5                    & 1.5                    & 1.5                     & 2                      & 2                      & 2                      \\
                & VIME           & 1.5                    & 1.5                    & 1.5                     & 1                      & 1                      & 1                      \\ \midrule
Robust              & TabTransf. & 1.5                    & 1.5                    & 1.5                     & 1.5                    & 1.5                    & 1.5                    \\
                & RLN            & 1.5                    & 1.5                    & 1.5                     & 2                      & 2                      & 2                      \\
                & VIME           & 1.5                    & 1.5                    & 1.5                     & 1                      & 1                      & 1                     \\ \bottomrule
\end{tabular}
    \caption{CTU - Rank of target robust accuracy when using different architectures as source with standard training (the lowest rank best for attacker), and as target (the highest rank the best for defender) }
    \label{tab:surrogate_ranking_ctu}
\end{table}

\subsection{Which attacker’s capabilities impact the success rate?}

\paragraph{Some architectures remain robust even with increased access to the target model.}
In the main paper, we evaluated the impact of the population size and number of search iterations. We evaluate below the impact of $\epsilon$ size distance and the number of search steps in gradient attack.

\textbf{Impact of the perturbation distance $\epsilon$.}

We present in tables \ref{tab:epsilon_CPGD}, \ref{tab:epsilon_CAPGD}, \ref{tab:epsilon_MOEVA}, and \ref{tab:epsilon_CAA} the impact of increasing the perturbation bound $\epsilon$ respectively for the CPGD attack, the CAPGD attack, the MOEVA attack, and the CAA attack.

\begin{table*}
\centering
\caption{Robust accuracy with different maximum perturbation $\epsilon$ for CPGD attack.}
\footnotesize
\label{tab:epsilon_CPGD}
\begin{tabular}{lll|lllllllll}
\toprule
        Dataset & & & \multicolumn{3}{c}{URL} & \multicolumn{3}{c}{LCLD} & \multicolumn{3}{c}{CTU} \\
        Model &    & &                    RLN &                  TabTrans. &                       VIME &                        RLN &                  TabTrans. &                       VIME &                        RLN &                  TabTrans. &                       VIME \\
Training & Cstr & $\epsilon$  &                            &                            &                            &                            &                            &                            &                            &                            &                            \\
\midrule
\multirow{6}{*}{Standard} & \multirow{3}{*}{Yes} & 0.25 &  $0.565$\tiny{$\pm 0.007$} &  $0.477$\tiny{$\pm 0.004$} &  $0.767$\tiny{$\pm 0.001$} &  $0.049$\tiny{$\pm 0.002$} &  $0.661$\tiny{$\pm 0.007$} &  $0.269$\tiny{$\pm 0.004$} &  $0.978$\tiny{$\pm 0.000$} &  $0.953$\tiny{$\pm 0.000$} &  $0.951$\tiny{$\pm 0.000$} \\
       &    & 0.5 &  $0.217$\tiny{$\pm 0.003$} &  $0.218$\tiny{$\pm 0.006$} &  $0.547$\tiny{$\pm 0.001$} &  $0.019$\tiny{$\pm 0.002$} &  $0.584$\tiny{$\pm 0.005$} &  $0.137$\tiny{$\pm 0.005$} &  $0.978$\tiny{$\pm 0.000$} &  $0.953$\tiny{$\pm 0.000$} &  $0.951$\tiny{$\pm 0.000$} \\
       &    & 1.0 &  $0.289$\tiny{$\pm 0.008$} &  $0.087$\tiny{$\pm 0.007$} &  $0.244$\tiny{$\pm 0.011$} &  $0.024$\tiny{$\pm 0.004$} &  $0.406$\tiny{$\pm 0.005$} &  $0.075$\tiny{$\pm 0.003$} &  $0.978$\tiny{$\pm 0.000$} &  $0.953$\tiny{$\pm 0.000$} &  $0.951$\tiny{$\pm 0.000$} \\
\cline{2-12}
       & \multirow{3}{*}{No} & 0.25 &  $0.742$\tiny{$\pm 0.001$} &  $0.492$\tiny{$\pm 0.001$} &  $0.776$\tiny{$\pm 0.001$} &  $0.681$\tiny{$\pm 0.000$} &  $0.706$\tiny{$\pm 0.000$} &  $0.666$\tiny{$\pm 0.000$} &  $0.978$\tiny{$\pm 0.000$} &  $0.953$\tiny{$\pm 0.000$} &  $0.951$\tiny{$\pm 0.000$} \\
       &    & 0.5 &  $0.904$\tiny{$\pm 0.001$} &  $0.818$\tiny{$\pm 0.005$} &  $0.570$\tiny{$\pm 0.003$} &  $0.681$\tiny{$\pm 0.000$} &  $0.706$\tiny{$\pm 0.000$} &  $0.666$\tiny{$\pm 0.000$} &  $0.978$\tiny{$\pm 0.000$} &  $0.953$\tiny{$\pm 0.000$} &  $0.951$\tiny{$\pm 0.000$} \\
       &    & 1.0 &  $0.879$\tiny{$\pm 0.005$} &  $0.826$\tiny{$\pm 0.010$} &  $0.442$\tiny{$\pm 0.007$} &  $0.681$\tiny{$\pm 0.000$} &  $0.706$\tiny{$\pm 0.000$} &  $0.666$\tiny{$\pm 0.000$} &  $0.978$\tiny{$\pm 0.000$} &  $0.953$\tiny{$\pm 0.000$} &  $0.951$\tiny{$\pm 0.000$} \\
\cline{1-12}
\cline{2-12}
\multirow{6}{*}{Robust} & \multirow{3}{*}{Yes} & 0.25 &  $0.920$\tiny{$\pm 0.001$} &  $0.922$\tiny{$\pm 0.000$} &  $0.895$\tiny{$\pm 0.001$} &  $0.694$\tiny{$\pm 0.003$} &  $0.736$\tiny{$\pm 0.001$} &  $0.363$\tiny{$\pm 0.004$} &  $0.973$\tiny{$\pm 0.000$} &  $0.953$\tiny{$\pm 0.000$} &  $0.951$\tiny{$\pm 0.000$} \\
       &    & 0.5 &  $0.811$\tiny{$\pm 0.002$} &  $0.812$\tiny{$\pm 0.003$} &  $0.789$\tiny{$\pm 0.002$} &  $0.692$\tiny{$\pm 0.001$} &  $0.729$\tiny{$\pm 0.003$} &  $0.253$\tiny{$\pm 0.004$} &  $0.973$\tiny{$\pm 0.000$} &  $0.953$\tiny{$\pm 0.000$} &  $0.951$\tiny{$\pm 0.000$} \\
       &    & 1.0 &  $0.290$\tiny{$\pm 0.005$} &  $0.437$\tiny{$\pm 0.003$} &  $0.347$\tiny{$\pm 0.006$} &  $0.688$\tiny{$\pm 0.002$} &  $0.725$\tiny{$\pm 0.001$} &  $0.183$\tiny{$\pm 0.010$} &  $0.973$\tiny{$\pm 0.000$} &  $0.953$\tiny{$\pm 0.000$} &  $0.951$\tiny{$\pm 0.000$} \\
\cline{2-12}
       & \multirow{3}{*}{No} & 0.25 &  $0.920$\tiny{$\pm 0.000$} &  $0.922$\tiny{$\pm 0.000$} &  $0.897$\tiny{$\pm 0.000$} &  $0.707$\tiny{$\pm 0.000$} &  $0.745$\tiny{$\pm 0.000$} &  $0.654$\tiny{$\pm 0.000$} &  $0.973$\tiny{$\pm 0.000$} &  $0.953$\tiny{$\pm 0.000$} &  $0.951$\tiny{$\pm 0.000$} \\
       &    & 0.5 &  $0.815$\tiny{$\pm 0.001$} &  $0.816$\tiny{$\pm 0.002$} &  $0.808$\tiny{$\pm 0.003$} &  $0.707$\tiny{$\pm 0.000$} &  $0.745$\tiny{$\pm 0.000$} &  $0.654$\tiny{$\pm 0.000$} &  $0.973$\tiny{$\pm 0.000$} &  $0.953$\tiny{$\pm 0.000$} &  $0.951$\tiny{$\pm 0.000$} \\
       &    & 1.0 &  $0.291$\tiny{$\pm 0.005$} &  $0.455$\tiny{$\pm 0.003$} &  $0.477$\tiny{$\pm 0.004$} &  $0.707$\tiny{$\pm 0.000$} &  $0.745$\tiny{$\pm 0.000$} &  $0.654$\tiny{$\pm 0.000$} &  $0.973$\tiny{$\pm 0.000$} &  $0.953$\tiny{$\pm 0.000$} &  $0.951$\tiny{$\pm 0.000$} \\
\bottomrule
\end{tabular}
\end{table*}

\begin{table*}
\centering
\caption{Robust accuracy with different maximum perturbation $\epsilon$ for CAPGD attack.}
\footnotesize
\label{tab:epsilon_CAPGD}
\begin{tabular}{lll|lllllllll}
\toprule
        Dataset &    && \multicolumn{3}{c}{URL} & \multicolumn{3}{c}{LCLD} & \multicolumn{3}{c}{CTU} \\
        Model&    & &                        RLN &                  TabTrans. &                       VIME &                        RLN &                  TabTrans. &                       VIME &                        RLN &                  TabTrans. &                       VIME \\
Training & Cstr & $\epsilon$ &                            &                            &                            &                            &                            &                            &                            &                            &                            \\
\midrule
\multirow{6}{*}{Standard} & \multirow{3}{*}{Yes} & 0.25 &  $0.518$\tiny{$\pm 0.006$} &  $0.451$\tiny{$\pm 0.003$} &  $0.767$\tiny{$\pm 0.002$} &  $0.299$\tiny{$\pm 0.005$} &  $0.565$\tiny{$\pm 0.007$} &  $0.490$\tiny{$\pm 0.005$} &  $0.978$\tiny{$\pm 0.000$} &  $0.953$\tiny{$\pm 0.000$} &  $0.951$\tiny{$\pm 0.000$} \\
       &    & 0.5 &  $0.266$\tiny{$\pm 0.012$} &  $0.193$\tiny{$\pm 0.006$} &  $0.582$\tiny{$\pm 0.001$} &  $0.265$\tiny{$\pm 0.007$} &  $0.477$\tiny{$\pm 0.009$} &  $0.369$\tiny{$\pm 0.012$} &  $0.978$\tiny{$\pm 0.000$} &  $0.953$\tiny{$\pm 0.000$} &  $0.951$\tiny{$\pm 0.000$} \\
       &    & 1.0 &  $0.208$\tiny{$\pm 0.012$} &  $0.048$\tiny{$\pm 0.003$} &  $0.302$\tiny{$\pm 0.004$} &  $0.245$\tiny{$\pm 0.012$} &  $0.355$\tiny{$\pm 0.016$} &  $0.298$\tiny{$\pm 0.012$} &  $0.978$\tiny{$\pm 0.000$} &  $0.953$\tiny{$\pm 0.000$} &  $0.951$\tiny{$\pm 0.000$} \\
\cline{2-12}
       & \multirow{3}{*}{No} & 0.25 &  $0.756$\tiny{$\pm 0.003$} &  $0.507$\tiny{$\pm 0.002$} &  $0.766$\tiny{$\pm 0.001$} &  $0.681$\tiny{$\pm 0.000$} &  $0.706$\tiny{$\pm 0.000$} &  $0.666$\tiny{$\pm 0.000$} &  $0.978$\tiny{$\pm 0.000$} &  $0.953$\tiny{$\pm 0.000$} &  $0.951$\tiny{$\pm 0.000$} \\
       &    & 0.5 &  $0.896$\tiny{$\pm 0.002$} &  $0.808$\tiny{$\pm 0.004$} &  $0.585$\tiny{$\pm 0.002$} &  $0.681$\tiny{$\pm 0.000$} &  $0.706$\tiny{$\pm 0.000$} &  $0.666$\tiny{$\pm 0.000$} &  $0.978$\tiny{$\pm 0.000$} &  $0.953$\tiny{$\pm 0.000$} &  $0.951$\tiny{$\pm 0.000$} \\
       &    & 1.0 &  $0.888$\tiny{$\pm 0.006$} &  $0.784$\tiny{$\pm 0.014$} &  $0.389$\tiny{$\pm 0.005$} &  $0.681$\tiny{$\pm 0.000$} &  $0.706$\tiny{$\pm 0.000$} &  $0.666$\tiny{$\pm 0.000$} &  $0.978$\tiny{$\pm 0.000$} &  $0.953$\tiny{$\pm 0.000$} &  $0.951$\tiny{$\pm 0.000$} \\
\cline{1-12}
\cline{2-12}
\multirow{6}{*}{Robust} & \multirow{3}{*}{Yes} & 0.25 &  $0.917$\tiny{$\pm 0.001$} &  $0.918$\tiny{$\pm 0.001$} &  $0.893$\tiny{$\pm 0.001$} &  $0.702$\tiny{$\pm 0.001$} &  $0.741$\tiny{$\pm 0.001$} &  $0.559$\tiny{$\pm 0.007$} &  $0.973$\tiny{$\pm 0.000$} &  $0.953$\tiny{$\pm 0.000$} &  $0.951$\tiny{$\pm 0.000$} \\
       &    & 0.5 &  $0.821$\tiny{$\pm 0.001$} &  $0.814$\tiny{$\pm 0.002$} &  $0.796$\tiny{$\pm 0.002$} &  $0.701$\tiny{$\pm 0.002$} &  $0.735$\tiny{$\pm 0.002$} &  $0.517$\tiny{$\pm 0.010$} &  $0.973$\tiny{$\pm 0.000$} &  $0.953$\tiny{$\pm 0.000$} &  $0.951$\tiny{$\pm 0.000$} \\
       &    & 1.0 &  $0.319$\tiny{$\pm 0.005$} &  $0.501$\tiny{$\pm 0.006$} &  $0.418$\tiny{$\pm 0.004$} &  $0.700$\tiny{$\pm 0.002$} &  $0.713$\tiny{$\pm 0.003$} &  $0.526$\tiny{$\pm 0.009$} &  $0.973$\tiny{$\pm 0.000$} &  $0.953$\tiny{$\pm 0.000$} &  $0.951$\tiny{$\pm 0.000$} \\
\cline{2-12}
       & \multirow{3}{*}{No} & 0.25 &  $0.917$\tiny{$\pm 0.001$} &  $0.917$\tiny{$\pm 0.000$} &  $0.895$\tiny{$\pm 0.000$} &  $0.707$\tiny{$\pm 0.000$} &  $0.745$\tiny{$\pm 0.000$} &  $0.654$\tiny{$\pm 0.000$} &  $0.973$\tiny{$\pm 0.000$} &  $0.953$\tiny{$\pm 0.000$} &  $0.951$\tiny{$\pm 0.000$} \\
       &    & 0.5 &  $0.817$\tiny{$\pm 0.002$} &  $0.813$\tiny{$\pm 0.002$} &  $0.805$\tiny{$\pm 0.003$} &  $0.707$\tiny{$\pm 0.000$} &  $0.745$\tiny{$\pm 0.000$} &  $0.654$\tiny{$\pm 0.000$} &  $0.973$\tiny{$\pm 0.000$} &  $0.953$\tiny{$\pm 0.000$} &  $0.951$\tiny{$\pm 0.000$} \\
       &    & 1.0 &  $0.284$\tiny{$\pm 0.004$} &  $0.504$\tiny{$\pm 0.004$} &  $0.423$\tiny{$\pm 0.005$} &  $0.707$\tiny{$\pm 0.000$} &  $0.745$\tiny{$\pm 0.000$} &  $0.654$\tiny{$\pm 0.000$} &  $0.973$\tiny{$\pm 0.000$} &  $0.953$\tiny{$\pm 0.000$} &  $0.951$\tiny{$\pm 0.000$} \\
\bottomrule
\end{tabular}
\end{table*}

\begin{table*}
\centering
\caption{Robust accuracy with different maximum perturbation $\epsilon$ for MOEVA attack.}
\footnotesize
\label{tab:epsilon_MOEVA}
\begin{tabular}{lll|lllllllll}
\toprule
       Dataset  &    & & \multicolumn{3}{c}{URL} & \multicolumn{3}{c}{LCLD} & \multicolumn{3}{c}{CTU} \\
        Model  &    & &                        RLN &                  TabTrans. &                       VIME &                        RLN &                  TabTrans. &                       VIME &                        RLN &                  TabTrans. &                       VIME \\
Training & Cstr & $\epsilon$ &                            &                            &                            &                            &                            &                            &                            &                            &                            \\
\midrule
\multirow{6}{*}{Standard} & \multirow{3}{*}{Yes} & 0.25 &  $0.372$\tiny{$\pm 0.005$} &  $0.316$\tiny{$\pm 0.002$} &  $0.701$\tiny{$\pm 0.003$} &  $0.013$\tiny{$\pm 0.003$} &  $0.275$\tiny{$\pm 0.004$} &  $0.365$\tiny{$\pm 0.007$} &  $0.952$\tiny{$\pm 0.003$} &  $0.953$\tiny{$\pm 0.000$} &  $0.561$\tiny{$\pm 0.034$} \\
       &    & 0.5 &  $0.236$\tiny{$\pm 0.005$} &  $0.182$\tiny{$\pm 0.008$} &  $0.565$\tiny{$\pm 0.009$} &  $0.006$\tiny{$\pm 0.002$} &  $0.150$\tiny{$\pm 0.007$} &  $0.269$\tiny{$\pm 0.018$} &  $0.940$\tiny{$\pm 0.002$} &  $0.953$\tiny{$\pm 0.000$} &  $0.408$\tiny{$\pm 0.047$} \\
       &    & 1.0 &  $0.227$\tiny{$\pm 0.006$} &  $0.172$\tiny{$\pm 0.008$} &  $0.391$\tiny{$\pm 0.034$} &  $0.006$\tiny{$\pm 0.002$} &  $0.133$\tiny{$\pm 0.009$} &  $0.262$\tiny{$\pm 0.018$} &  $0.940$\tiny{$\pm 0.002$} &  $0.951$\tiny{$\pm 0.002$} &  $0.340$\tiny{$\pm 0.040$} \\
\cline{2-12}
       & \multirow{3}{*}{No} & 0.25 &  $0.379$\tiny{$\pm 0.004$} &  $0.317$\tiny{$\pm 0.004$} &  $0.699$\tiny{$\pm 0.006$} &  $0.542$\tiny{$\pm 0.008$} &  $0.671$\tiny{$\pm 0.002$} &  $0.655$\tiny{$\pm 0.002$} &  $0.976$\tiny{$\pm 0.001$} &  $0.953$\tiny{$\pm 0.000$} &  $0.950$\tiny{$\pm 0.001$} \\
       &    & 0.5 &  $0.243$\tiny{$\pm 0.004$} &  $0.184$\tiny{$\pm 0.008$} &  $0.561$\tiny{$\pm 0.009$} &  $0.542$\tiny{$\pm 0.008$} &  $0.671$\tiny{$\pm 0.002$} &  $0.655$\tiny{$\pm 0.002$} &  $0.974$\tiny{$\pm 0.002$} &  $0.953$\tiny{$\pm 0.000$} &  $0.950$\tiny{$\pm 0.001$} \\
       &    & 1.0 &  $0.233$\tiny{$\pm 0.006$} &  $0.175$\tiny{$\pm 0.008$} &  $0.386$\tiny{$\pm 0.036$} &  $0.542$\tiny{$\pm 0.008$} &  $0.671$\tiny{$\pm 0.002$} &  $0.655$\tiny{$\pm 0.002$} &  $0.974$\tiny{$\pm 0.002$} &  $0.953$\tiny{$\pm 0.000$} &  $0.950$\tiny{$\pm 0.001$} \\
\cline{1-12}
\cline{2-12}
\multirow{6}{*}{Robust} & \multirow{3}{*}{Yes} & 0.25 &  $0.832$\tiny{$\pm 0.002$} &  $0.830$\tiny{$\pm 0.002$} &  $0.827$\tiny{$\pm 0.002$} &  $0.673$\tiny{$\pm 0.001$} &  $0.720$\tiny{$\pm 0.001$} &  $0.358$\tiny{$\pm 0.006$} &  $0.971$\tiny{$\pm 0.001$} &  $0.953$\tiny{$\pm 0.000$} &  $0.951$\tiny{$\pm 0.000$} \\
       &    & 0.5 &  $0.563$\tiny{$\pm 0.007$} &  $0.568$\tiny{$\pm 0.008$} &  $0.700$\tiny{$\pm 0.006$} &  $0.662$\tiny{$\pm 0.002$} &  $0.713$\tiny{$\pm 0.002$} &  $0.226$\tiny{$\pm 0.008$} &  $0.971$\tiny{$\pm 0.000$} &  $0.953$\tiny{$\pm 0.000$} &  $0.940$\tiny{$\pm 0.005$} \\
       &    & 1.0 &  $0.349$\tiny{$\pm 0.035$} &  $0.328$\tiny{$\pm 0.046$} &  $0.418$\tiny{$\pm 0.046$} &  $0.660$\tiny{$\pm 0.003$} &  $0.710$\tiny{$\pm 0.002$} &  $0.205$\tiny{$\pm 0.008$} &  $0.971$\tiny{$\pm 0.000$} &  $0.952$\tiny{$\pm 0.002$} &  $0.910$\tiny{$\pm 0.017$} \\
\cline{2-12}
       & \multirow{3}{*}{No} & 0.25 &  $0.831$\tiny{$\pm 0.002$} &  $0.832$\tiny{$\pm 0.001$} &  $0.827$\tiny{$\pm 0.002$} &  $0.691$\tiny{$\pm 0.001$} &  $0.744$\tiny{$\pm 0.000$} &  $0.626$\tiny{$\pm 0.003$} &  $0.972$\tiny{$\pm 0.001$} &  $0.953$\tiny{$\pm 0.000$} &  $0.951$\tiny{$\pm 0.000$} \\
       &    & 0.5 &  $0.560$\tiny{$\pm 0.005$} &  $0.571$\tiny{$\pm 0.008$} &  $0.702$\tiny{$\pm 0.005$} &  $0.690$\tiny{$\pm 0.002$} &  $0.743$\tiny{$\pm 0.001$} &  $0.623$\tiny{$\pm 0.003$} &  $0.972$\tiny{$\pm 0.001$} &  $0.953$\tiny{$\pm 0.000$} &  $0.942$\tiny{$\pm 0.006$} \\
       &    & 1.0 &  $0.347$\tiny{$\pm 0.034$} &  $0.333$\tiny{$\pm 0.043$} &  $0.421$\tiny{$\pm 0.046$} &  $0.690$\tiny{$\pm 0.002$} &  $0.743$\tiny{$\pm 0.001$} &  $0.622$\tiny{$\pm 0.002$} &  $0.972$\tiny{$\pm 0.001$} &  $0.953$\tiny{$\pm 0.000$} &  $0.921$\tiny{$\pm 0.015$} \\
\bottomrule
\end{tabular}
\end{table*}

\begin{table*}
\centering
\caption{Robust accuracy with different maximum perturbation $\epsilon$ for CAA attack.}
\footnotesize
\label{tab:epsilon_CAA}
\begin{tabular}{lll|lllllllll}
\toprule
        Dataset&    & & \multicolumn{3}{c}{URL} & \multicolumn{3}{c}{LCLD} & \multicolumn{3}{c}{CTU} \\
        Model&    & &                        RLN &                  TabTrans. &                       VIME &                        RLN &                  TabTrans. &                       VIME &                        RLN &                  TabTrans. &                       VIME \\
Training & Cstr & $\epsilon$ &                            &                            &                            &                            &                            &                            &                            &                            &                            \\
\midrule
\multirow{6}{*}{Standard} & \multirow{3}{*}{Yes} & 0.25 &  $0.366$\tiny{$\pm 0.004$} &  $0.315$\tiny{$\pm 0.001$} &  $0.697$\tiny{$\pm 0.003$} &  $0.002$\tiny{$\pm 0.000$} &  $0.273$\tiny{$\pm 0.003$} &  $0.224$\tiny{$\pm 0.002$} &  $0.952$\tiny{$\pm 0.003$} &  $0.953$\tiny{$\pm 0.000$} &  $0.561$\tiny{$\pm 0.034$} \\
       &    & 0.5 &  $0.123$\tiny{$\pm 0.006$} &  $0.088$\tiny{$\pm 0.002$} &  $0.492$\tiny{$\pm 0.005$} &  $0.001$\tiny{$\pm 0.001$} &  $0.135$\tiny{$\pm 0.006$} &  $0.066$\tiny{$\pm 0.005$} &  $0.940$\tiny{$\pm 0.002$} &  $0.953$\tiny{$\pm 0.000$} &  $0.408$\tiny{$\pm 0.047$} \\
       &    & 1.0 &  $0.115$\tiny{$\pm 0.006$} &  $0.014$\tiny{$\pm 0.001$} &  $0.161$\tiny{$\pm 0.004$} &  $0.000$\tiny{$\pm 0.000$} &  $0.080$\tiny{$\pm 0.005$} &  $0.022$\tiny{$\pm 0.004$} &  $0.940$\tiny{$\pm 0.002$} &  $0.951$\tiny{$\pm 0.002$} &  $0.340$\tiny{$\pm 0.040$} \\
\cline{2-12}
       & \multirow{3}{*}{No} & 0.25 &  $0.371$\tiny{$\pm 0.003$} &  $0.316$\tiny{$\pm 0.004$} &  $0.696$\tiny{$\pm 0.005$} &  $0.542$\tiny{$\pm 0.008$} &  $0.671$\tiny{$\pm 0.002$} &  $0.655$\tiny{$\pm 0.002$} &  $0.976$\tiny{$\pm 0.001$} &  $0.953$\tiny{$\pm 0.000$} &  $0.950$\tiny{$\pm 0.001$} \\
       &    & 0.5 &  $0.224$\tiny{$\pm 0.003$} &  $0.131$\tiny{$\pm 0.006$} &  $0.499$\tiny{$\pm 0.004$} &  $0.542$\tiny{$\pm 0.008$} &  $0.671$\tiny{$\pm 0.002$} &  $0.655$\tiny{$\pm 0.002$} &  $0.974$\tiny{$\pm 0.002$} &  $0.953$\tiny{$\pm 0.000$} &  $0.950$\tiny{$\pm 0.001$} \\
       &    & 1.0 &  $0.220$\tiny{$\pm 0.005$} &  $0.142$\tiny{$\pm 0.008$} &  $0.168$\tiny{$\pm 0.007$} &  $0.542$\tiny{$\pm 0.008$} &  $0.671$\tiny{$\pm 0.002$} &  $0.655$\tiny{$\pm 0.002$} &  $0.974$\tiny{$\pm 0.002$} &  $0.953$\tiny{$\pm 0.000$} &  $0.950$\tiny{$\pm 0.001$} \\
\cline{1-12}
\cline{2-12}
\multirow{6}{*}{Robust} & \multirow{3}{*}{Yes} & 0.25 &  $0.832$\tiny{$\pm 0.002$} &  $0.830$\tiny{$\pm 0.002$} &  $0.827$\tiny{$\pm 0.002$} &  $0.672$\tiny{$\pm 0.002$} &  $0.719$\tiny{$\pm 0.002$} &  $0.302$\tiny{$\pm 0.003$} &  $0.971$\tiny{$\pm 0.001$} &  $0.953$\tiny{$\pm 0.000$} &  $0.951$\tiny{$\pm 0.000$} \\
       &    & 0.5 &  $0.562$\tiny{$\pm 0.006$} &  $0.567$\tiny{$\pm 0.008$} &  $0.697$\tiny{$\pm 0.006$} &  $0.661$\tiny{$\pm 0.003$} &  $0.708$\tiny{$\pm 0.002$} &  $0.147$\tiny{$\pm 0.005$} &  $0.971$\tiny{$\pm 0.000$} &  $0.953$\tiny{$\pm 0.000$} &  $0.940$\tiny{$\pm 0.005$} \\
       &    & 1.0 &  $0.223$\tiny{$\pm 0.009$} &  $0.252$\tiny{$\pm 0.022$} &  $0.272$\tiny{$\pm 0.008$} &  $0.656$\tiny{$\pm 0.003$} &  $0.692$\tiny{$\pm 0.003$} &  $0.104$\tiny{$\pm 0.003$} &  $0.971$\tiny{$\pm 0.000$} &  $0.952$\tiny{$\pm 0.002$} &  $0.910$\tiny{$\pm 0.017$} \\
\cline{2-12}
       & \multirow{3}{*}{No} & 0.25 &  $0.831$\tiny{$\pm 0.002$} &  $0.832$\tiny{$\pm 0.001$} &  $0.827$\tiny{$\pm 0.002$} &  $0.691$\tiny{$\pm 0.001$} &  $0.744$\tiny{$\pm 0.000$} &  $0.626$\tiny{$\pm 0.003$} &  $0.972$\tiny{$\pm 0.001$} &  $0.953$\tiny{$\pm 0.000$} &  $0.951$\tiny{$\pm 0.000$} \\
       &    & 0.5 &  $0.559$\tiny{$\pm 0.005$} &  $0.570$\tiny{$\pm 0.008$} &  $0.699$\tiny{$\pm 0.003$} &  $0.690$\tiny{$\pm 0.002$} &  $0.743$\tiny{$\pm 0.001$} &  $0.623$\tiny{$\pm 0.003$} &  $0.972$\tiny{$\pm 0.001$} &  $0.953$\tiny{$\pm 0.000$} &  $0.942$\tiny{$\pm 0.006$} \\
       &    & 1.0 &  $0.213$\tiny{$\pm 0.006$} &  $0.258$\tiny{$\pm 0.026$} &  $0.282$\tiny{$\pm 0.010$} &  $0.690$\tiny{$\pm 0.002$} &  $0.743$\tiny{$\pm 0.001$} &  $0.622$\tiny{$\pm 0.002$} &  $0.972$\tiny{$\pm 0.001$} &  $0.953$\tiny{$\pm 0.000$} &  $0.921$\tiny{$\pm 0.015$} \\
\bottomrule
\end{tabular}
\end{table*}

\textbf{Impact of the number of steps.}

We present in tables \ref{tab:step_CPGD}, \ref{tab:step_CAPGD}, and \ref{tab:step_CAA} the impact of increasing the number of perturbation steps respectively for the CPGD attack, the CAPGD attack, and the CAA attack.

\begin{table*}
\centering
\caption{Robust accuracy with different \#step for CPGD attack.}
\footnotesize
\label{tab:step_CPGD}
\begin{tabular}{lll|lllllllll}
\toprule
        Dataset &    & & \multicolumn{3}{c}{URL} & \multicolumn{3}{c}{LCLD} & \multicolumn{3}{c}{CTU} \\
        Model &    & &                        RLN &                  TabTrans. &                       VIME &                        RLN &                  TabTrans. &                       VIME &                        RLN &                  TabTrans. &                       VIME \\
Training & Cstr & Stp &                            &                            &                            &                            &                            &                            &                            &                            &                            \\
\midrule
\multirow{8}{*}{Standard} & \multirow{4}{*}{Yes} & 10  &  $0.217$\tiny{$\pm 0.003$} &  $0.218$\tiny{$\pm 0.006$} &  $0.547$\tiny{$\pm 0.001$} &  $0.019$\tiny{$\pm 0.002$} &  $0.584$\tiny{$\pm 0.005$} &  $0.137$\tiny{$\pm 0.005$} &  $0.978$\tiny{$\pm 0.000$} &  $0.953$\tiny{$\pm 0.000$} &  $0.951$\tiny{$\pm 0.000$} \\
       &    & 20  &  $0.238$\tiny{$\pm 0.007$} &  $0.168$\tiny{$\pm 0.006$} &  $0.544$\tiny{$\pm 0.002$} &  $0.012$\tiny{$\pm 0.003$} &  $0.574$\tiny{$\pm 0.008$} &  $0.114$\tiny{$\pm 0.004$} &  $0.978$\tiny{$\pm 0.000$} &  $0.953$\tiny{$\pm 0.000$} &  $0.951$\tiny{$\pm 0.000$} \\
       &    & 50  &  $0.190$\tiny{$\pm 0.004$} &  $0.167$\tiny{$\pm 0.007$} &  $0.547$\tiny{$\pm 0.001$} &  $0.010$\tiny{$\pm 0.001$} &  $0.575$\tiny{$\pm 0.007$} &  $0.100$\tiny{$\pm 0.011$} &  $0.978$\tiny{$\pm 0.000$} &  $0.953$\tiny{$\pm 0.000$} &  $0.951$\tiny{$\pm 0.000$} \\
       &    & 100 &  $0.195$\tiny{$\pm 0.006$} &  $0.172$\tiny{$\pm 0.005$} &  $0.546$\tiny{$\pm 0.001$} &  $0.008$\tiny{$\pm 0.002$} &  $0.576$\tiny{$\pm 0.002$} &  $0.092$\tiny{$\pm 0.005$} &  $0.978$\tiny{$\pm 0.000$} &  $0.953$\tiny{$\pm 0.000$} &  $0.951$\tiny{$\pm 0.000$} \\
\cline{2-12}
       & \multirow{4}{*}{No} & 10  &  $0.904$\tiny{$\pm 0.001$} &  $0.818$\tiny{$\pm 0.005$} &  $0.570$\tiny{$\pm 0.003$} &  $0.681$\tiny{$\pm 0.000$} &  $0.706$\tiny{$\pm 0.000$} &  $0.666$\tiny{$\pm 0.000$} &  $0.978$\tiny{$\pm 0.000$} &  $0.953$\tiny{$\pm 0.000$} &  $0.951$\tiny{$\pm 0.000$} \\
       &    & 20  &  $0.911$\tiny{$\pm 0.003$} &  $0.825$\tiny{$\pm 0.003$} &  $0.569$\tiny{$\pm 0.003$} &  $0.681$\tiny{$\pm 0.000$} &  $0.706$\tiny{$\pm 0.000$} &  $0.666$\tiny{$\pm 0.000$} &  $0.978$\tiny{$\pm 0.000$} &  $0.953$\tiny{$\pm 0.000$} &  $0.951$\tiny{$\pm 0.000$} \\
       &    & 50  &  $0.910$\tiny{$\pm 0.003$} &  $0.824$\tiny{$\pm 0.002$} &  $0.569$\tiny{$\pm 0.001$} &  $0.681$\tiny{$\pm 0.000$} &  $0.706$\tiny{$\pm 0.000$} &  $0.666$\tiny{$\pm 0.000$} &  $0.978$\tiny{$\pm 0.000$} &  $0.953$\tiny{$\pm 0.000$} &  $0.951$\tiny{$\pm 0.000$} \\
       &    & 100 &  $0.908$\tiny{$\pm 0.002$} &  $0.824$\tiny{$\pm 0.002$} &  $0.568$\tiny{$\pm 0.001$} &  $0.681$\tiny{$\pm 0.000$} &  $0.706$\tiny{$\pm 0.000$} &  $0.666$\tiny{$\pm 0.000$} &  $0.978$\tiny{$\pm 0.000$} &  $0.953$\tiny{$\pm 0.000$} &  $0.951$\tiny{$\pm 0.000$} \\
\cline{1-12}
\cline{2-12}
\multirow{8}{*}{Robust} & \multirow{4}{*}{Yes} & 10  &  $0.811$\tiny{$\pm 0.002$} &  $0.812$\tiny{$\pm 0.003$} &  $0.789$\tiny{$\pm 0.002$} &  $0.692$\tiny{$\pm 0.001$} &  $0.729$\tiny{$\pm 0.003$} &  $0.253$\tiny{$\pm 0.004$} &  $0.973$\tiny{$\pm 0.000$} &  $0.953$\tiny{$\pm 0.000$} &  $0.951$\tiny{$\pm 0.000$} \\
       &    & 20  &  $0.812$\tiny{$\pm 0.001$} &  $0.819$\tiny{$\pm 0.001$} &  $0.788$\tiny{$\pm 0.002$} &  $0.685$\tiny{$\pm 0.004$} &  $0.721$\tiny{$\pm 0.003$} &  $0.238$\tiny{$\pm 0.007$} &  $0.973$\tiny{$\pm 0.000$} &  $0.953$\tiny{$\pm 0.000$} &  $0.951$\tiny{$\pm 0.000$} \\
       &    & 50  &  $0.812$\tiny{$\pm 0.000$} &  $0.820$\tiny{$\pm 0.001$} &  $0.788$\tiny{$\pm 0.001$} &  $0.679$\tiny{$\pm 0.003$} &  $0.718$\tiny{$\pm 0.003$} &  $0.244$\tiny{$\pm 0.004$} &  $0.973$\tiny{$\pm 0.000$} &  $0.953$\tiny{$\pm 0.000$} &  $0.951$\tiny{$\pm 0.000$} \\
       &    & 100 &  $0.812$\tiny{$\pm 0.000$} &  $0.820$\tiny{$\pm 0.001$} &  $0.789$\tiny{$\pm 0.001$} &  $0.675$\tiny{$\pm 0.002$} &  $0.718$\tiny{$\pm 0.003$} &  $0.232$\tiny{$\pm 0.008$} &  $0.973$\tiny{$\pm 0.000$} &  $0.953$\tiny{$\pm 0.000$} &  $0.951$\tiny{$\pm 0.000$} \\
\cline{2-12}
       & \multirow{4}{*}{No} & 10  &  $0.815$\tiny{$\pm 0.001$} &  $0.816$\tiny{$\pm 0.002$} &  $0.808$\tiny{$\pm 0.003$} &  $0.707$\tiny{$\pm 0.000$} &  $0.745$\tiny{$\pm 0.000$} &  $0.654$\tiny{$\pm 0.000$} &  $0.973$\tiny{$\pm 0.000$} &  $0.953$\tiny{$\pm 0.000$} &  $0.951$\tiny{$\pm 0.000$} \\
       &    & 20  &  $0.812$\tiny{$\pm 0.001$} &  $0.821$\tiny{$\pm 0.001$} &  $0.804$\tiny{$\pm 0.001$} &  $0.707$\tiny{$\pm 0.000$} &  $0.745$\tiny{$\pm 0.000$} &  $0.654$\tiny{$\pm 0.000$} &  $0.973$\tiny{$\pm 0.000$} &  $0.953$\tiny{$\pm 0.000$} &  $0.951$\tiny{$\pm 0.000$} \\
       &    & 50  &  $0.811$\tiny{$\pm 0.000$} &  $0.822$\tiny{$\pm 0.001$} &  $0.803$\tiny{$\pm 0.002$} &  $0.707$\tiny{$\pm 0.000$} &  $0.745$\tiny{$\pm 0.000$} &  $0.654$\tiny{$\pm 0.000$} &  $0.973$\tiny{$\pm 0.000$} &  $0.953$\tiny{$\pm 0.000$} &  $0.951$\tiny{$\pm 0.000$} \\
       &    & 100 &  $0.812$\tiny{$\pm 0.000$} &  $0.820$\tiny{$\pm 0.002$} &  $0.803$\tiny{$\pm 0.002$} &  $0.707$\tiny{$\pm 0.000$} &  $0.745$\tiny{$\pm 0.000$} &  $0.654$\tiny{$\pm 0.000$} &  $0.973$\tiny{$\pm 0.000$} &  $0.953$\tiny{$\pm 0.000$} &  $0.951$\tiny{$\pm 0.000$} \\
\bottomrule
\end{tabular}
\end{table*}

\begin{table*}
\centering
\caption{Robust accuracy with different \#step for CAPGD attack.}
\footnotesize
\label{tab:step_CAPGD}
\begin{tabular}{lll|lllllllll}
\toprule
       Dataset &    & & \multicolumn{3}{c}{URL} & \multicolumn{3}{c}{LCLD} & \multicolumn{3}{c}{CTU} \\
       Model &    & &                        RLN &                  TabTrans. &                       VIME &                        RLN &                  TabTrans. &                       VIME &                        RLN &                  TabTrans. &                       VIME \\
Training & Cstr & Stp &                            &                            &                            &                            &                            &                            &                            &                            &                            \\
\midrule
\multirow{8}{*}{Standard} & \multirow{4}{*}{Yes} & 10  &  $0.266$\tiny{$\pm 0.012$} &  $0.193$\tiny{$\pm 0.006$} &  $0.582$\tiny{$\pm 0.001$} &  $0.265$\tiny{$\pm 0.007$} &  $0.477$\tiny{$\pm 0.009$} &  $0.369$\tiny{$\pm 0.012$} &  $0.978$\tiny{$\pm 0.000$} &  $0.953$\tiny{$\pm 0.000$} &  $0.951$\tiny{$\pm 0.000$} \\
       &    & 20  &  $0.093$\tiny{$\pm 0.005$} &  $0.087$\tiny{$\pm 0.003$} &  $0.508$\tiny{$\pm 0.003$} &  $0.185$\tiny{$\pm 0.005$} &  $0.428$\tiny{$\pm 0.005$} &  $0.332$\tiny{$\pm 0.011$} &  $0.978$\tiny{$\pm 0.000$} &  $0.953$\tiny{$\pm 0.000$} &  $0.951$\tiny{$\pm 0.000$} \\
       &    & 50  &  $0.117$\tiny{$\pm 0.003$} &  $0.113$\tiny{$\pm 0.003$} &  $0.490$\tiny{$\pm 0.002$} &  $0.019$\tiny{$\pm 0.003$} &  $0.256$\tiny{$\pm 0.004$} &  $0.161$\tiny{$\pm 0.005$} &  $0.978$\tiny{$\pm 0.000$} &  $0.953$\tiny{$\pm 0.000$} &  $0.951$\tiny{$\pm 0.000$} \\
       &    & 100 &  $0.116$\tiny{$\pm 0.004$} &  $0.110$\tiny{$\pm 0.002$} &  $0.493$\tiny{$\pm 0.002$} &  $0.003$\tiny{$\pm 0.001$} &  $0.193$\tiny{$\pm 0.001$} &  $0.100$\tiny{$\pm 0.003$} &  $0.978$\tiny{$\pm 0.000$} &  $0.953$\tiny{$\pm 0.000$} &  $0.951$\tiny{$\pm 0.000$} \\
\cline{2-12}
       & \multirow{4}{*}{No} & 10  &  $0.896$\tiny{$\pm 0.002$} &  $0.808$\tiny{$\pm 0.004$} &  $0.585$\tiny{$\pm 0.002$} &  $0.681$\tiny{$\pm 0.000$} &  $0.706$\tiny{$\pm 0.000$} &  $0.666$\tiny{$\pm 0.000$} &  $0.978$\tiny{$\pm 0.000$} &  $0.953$\tiny{$\pm 0.000$} &  $0.951$\tiny{$\pm 0.000$} \\
       &    & 20  &  $0.892$\tiny{$\pm 0.001$} &  $0.810$\tiny{$\pm 0.006$} &  $0.530$\tiny{$\pm 0.003$} &  $0.681$\tiny{$\pm 0.000$} &  $0.706$\tiny{$\pm 0.000$} &  $0.666$\tiny{$\pm 0.000$} &  $0.978$\tiny{$\pm 0.000$} &  $0.953$\tiny{$\pm 0.000$} &  $0.951$\tiny{$\pm 0.000$} \\
       &    & 50  &  $0.902$\tiny{$\pm 0.001$} &  $0.831$\tiny{$\pm 0.003$} &  $0.524$\tiny{$\pm 0.002$} &  $0.681$\tiny{$\pm 0.000$} &  $0.706$\tiny{$\pm 0.000$} &  $0.666$\tiny{$\pm 0.000$} &  $0.978$\tiny{$\pm 0.000$} &  $0.953$\tiny{$\pm 0.000$} &  $0.951$\tiny{$\pm 0.000$} \\
       &    & 100 &  $0.892$\tiny{$\pm 0.001$} &  $0.820$\tiny{$\pm 0.005$} &  $0.518$\tiny{$\pm 0.003$} &  $0.681$\tiny{$\pm 0.000$} &  $0.706$\tiny{$\pm 0.000$} &  $0.666$\tiny{$\pm 0.000$} &  $0.978$\tiny{$\pm 0.000$} &  $0.953$\tiny{$\pm 0.000$} &  $0.951$\tiny{$\pm 0.000$} \\
\cline{1-12}
\cline{2-12}
\multirow{8}{*}{Robust} & \multirow{4}{*}{Yes} & 10  &  $0.821$\tiny{$\pm 0.001$} &  $0.814$\tiny{$\pm 0.002$} &  $0.796$\tiny{$\pm 0.002$} &  $0.701$\tiny{$\pm 0.002$} &  $0.735$\tiny{$\pm 0.002$} &  $0.517$\tiny{$\pm 0.010$} &  $0.973$\tiny{$\pm 0.000$} &  $0.953$\tiny{$\pm 0.000$} &  $0.951$\tiny{$\pm 0.000$} \\
       &    & 20  &  $0.805$\tiny{$\pm 0.001$} &  $0.808$\tiny{$\pm 0.002$} &  $0.774$\tiny{$\pm 0.001$} &  $0.700$\tiny{$\pm 0.002$} &  $0.737$\tiny{$\pm 0.001$} &  $0.498$\tiny{$\pm 0.009$} &  $0.973$\tiny{$\pm 0.000$} &  $0.953$\tiny{$\pm 0.000$} &  $0.951$\tiny{$\pm 0.000$} \\
       &    & 50  &  $0.804$\tiny{$\pm 0.001$} &  $0.811$\tiny{$\pm 0.000$} &  $0.779$\tiny{$\pm 0.002$} &  $0.699$\tiny{$\pm 0.002$} &  $0.734$\tiny{$\pm 0.002$} &  $0.418$\tiny{$\pm 0.007$} &  $0.973$\tiny{$\pm 0.000$} &  $0.953$\tiny{$\pm 0.000$} &  $0.951$\tiny{$\pm 0.000$} \\
       &    & 100 &  $0.801$\tiny{$\pm 0.001$} &  $0.810$\tiny{$\pm 0.001$} &  $0.773$\tiny{$\pm 0.002$} &  $0.698$\tiny{$\pm 0.002$} &  $0.731$\tiny{$\pm 0.001$} &  $0.387$\tiny{$\pm 0.004$} &  $0.973$\tiny{$\pm 0.000$} &  $0.953$\tiny{$\pm 0.000$} &  $0.951$\tiny{$\pm 0.000$} \\
\cline{2-12}
       & \multirow{4}{*}{No} & 10  &  $0.817$\tiny{$\pm 0.002$} &  $0.813$\tiny{$\pm 0.002$} &  $0.805$\tiny{$\pm 0.003$} &  $0.707$\tiny{$\pm 0.000$} &  $0.745$\tiny{$\pm 0.000$} &  $0.654$\tiny{$\pm 0.000$} &  $0.973$\tiny{$\pm 0.000$} &  $0.953$\tiny{$\pm 0.000$} &  $0.951$\tiny{$\pm 0.000$} \\
       &    & 20  &  $0.805$\tiny{$\pm 0.002$} &  $0.810$\tiny{$\pm 0.001$} &  $0.792$\tiny{$\pm 0.001$} &  $0.707$\tiny{$\pm 0.000$} &  $0.745$\tiny{$\pm 0.000$} &  $0.654$\tiny{$\pm 0.000$} &  $0.973$\tiny{$\pm 0.000$} &  $0.953$\tiny{$\pm 0.000$} &  $0.951$\tiny{$\pm 0.000$} \\
       &    & 50  &  $0.804$\tiny{$\pm 0.001$} &  $0.813$\tiny{$\pm 0.001$} &  $0.790$\tiny{$\pm 0.002$} &  $0.707$\tiny{$\pm 0.000$} &  $0.745$\tiny{$\pm 0.000$} &  $0.654$\tiny{$\pm 0.000$} &  $0.973$\tiny{$\pm 0.000$} &  $0.953$\tiny{$\pm 0.000$} &  $0.951$\tiny{$\pm 0.000$} \\
       &    & 100 &  $0.803$\tiny{$\pm 0.001$} &  $0.813$\tiny{$\pm 0.000$} &  $0.788$\tiny{$\pm 0.002$} &  $0.707$\tiny{$\pm 0.000$} &  $0.745$\tiny{$\pm 0.000$} &  $0.654$\tiny{$\pm 0.000$} &  $0.973$\tiny{$\pm 0.000$} &  $0.953$\tiny{$\pm 0.000$} &  $0.951$\tiny{$\pm 0.000$} \\
\bottomrule
\end{tabular}
\end{table*}

\begin{table*}
\centering
\caption{Robust accuracy with different \#step for CAA attack.}
\footnotesize
\label{tab:step_CAA}
\begin{tabular}{lll|lllllllll}
\toprule
        Dataset&    & & \multicolumn{3}{c}{URL} & \multicolumn{3}{c}{LCLD} & \multicolumn{3}{c}{CTU} \\
         Model&    & &                        RLN &                  TabTrans. &                       VIME &                        RLN &                  TabTrans. &                       VIME &                        RLN &                  TabTrans. &                       VIME \\
Training & Cstr & Stp &                            &                            &                            &                            &                            &                            &                            &                            &                            \\
\midrule
\multirow{8}{*}{Standard} & \multirow{4}{*}{Yes} & 10  &  $0.123$\tiny{$\pm 0.006$} &  $0.088$\tiny{$\pm 0.002$} &  $0.492$\tiny{$\pm 0.005$} &  $0.001$\tiny{$\pm 0.001$} &  $0.135$\tiny{$\pm 0.006$} &  $0.066$\tiny{$\pm 0.005$} &  $0.940$\tiny{$\pm 0.002$} &  $0.953$\tiny{$\pm 0.000$} &  $0.408$\tiny{$\pm 0.047$} \\
       &    & 20  &  $0.086$\tiny{$\pm 0.003$} &  $0.079$\tiny{$\pm 0.000$} &  $0.470$\tiny{$\pm 0.003$} &  $0.000$\tiny{$\pm 0.000$} &  $0.132$\tiny{$\pm 0.003$} &  $0.056$\tiny{$\pm 0.003$} &  $0.940$\tiny{$\pm 0.002$} &  $0.953$\tiny{$\pm 0.000$} &  $0.408$\tiny{$\pm 0.047$} \\
       &    & 50  &  $0.091$\tiny{$\pm 0.003$} &  $0.087$\tiny{$\pm 0.002$} &  $0.463$\tiny{$\pm 0.003$} &  $0.000$\tiny{$\pm 0.000$} &  $0.120$\tiny{$\pm 0.005$} &  $0.042$\tiny{$\pm 0.004$} &  $0.940$\tiny{$\pm 0.002$} &  $0.953$\tiny{$\pm 0.000$} &  $0.408$\tiny{$\pm 0.047$} \\
       &    & 100 &  $0.087$\tiny{$\pm 0.002$} &  $0.084$\tiny{$\pm 0.001$} &  $0.459$\tiny{$\pm 0.002$} &  $0.000$\tiny{$\pm 0.000$} &  $0.114$\tiny{$\pm 0.005$} &  $0.033$\tiny{$\pm 0.001$} &  $0.940$\tiny{$\pm 0.002$} &  $0.953$\tiny{$\pm 0.000$} &  $0.408$\tiny{$\pm 0.047$} \\
\cline{2-12}
       & \multirow{4}{*}{No} & 10  &  $0.224$\tiny{$\pm 0.003$} &  $0.131$\tiny{$\pm 0.006$} &  $0.499$\tiny{$\pm 0.004$} &  $0.542$\tiny{$\pm 0.008$} &  $0.671$\tiny{$\pm 0.002$} &  $0.655$\tiny{$\pm 0.002$} &  $0.974$\tiny{$\pm 0.002$} &  $0.953$\tiny{$\pm 0.000$} &  $0.950$\tiny{$\pm 0.001$} \\
       &    & 20  &  $0.224$\tiny{$\pm 0.005$} &  $0.132$\tiny{$\pm 0.006$} &  $0.479$\tiny{$\pm 0.003$} &  $0.542$\tiny{$\pm 0.008$} &  $0.671$\tiny{$\pm 0.002$} &  $0.655$\tiny{$\pm 0.002$} &  $0.974$\tiny{$\pm 0.002$} &  $0.953$\tiny{$\pm 0.000$} &  $0.950$\tiny{$\pm 0.001$} \\
       &    & 50  &  $0.223$\tiny{$\pm 0.004$} &  $0.134$\tiny{$\pm 0.007$} &  $0.474$\tiny{$\pm 0.003$} &  $0.542$\tiny{$\pm 0.008$} &  $0.671$\tiny{$\pm 0.002$} &  $0.655$\tiny{$\pm 0.002$} &  $0.974$\tiny{$\pm 0.002$} &  $0.953$\tiny{$\pm 0.000$} &  $0.950$\tiny{$\pm 0.001$} \\
       &    & 100 &  $0.222$\tiny{$\pm 0.005$} &  $0.132$\tiny{$\pm 0.007$} &  $0.469$\tiny{$\pm 0.004$} &  $0.542$\tiny{$\pm 0.008$} &  $0.671$\tiny{$\pm 0.002$} &  $0.655$\tiny{$\pm 0.002$} &  $0.974$\tiny{$\pm 0.002$} &  $0.953$\tiny{$\pm 0.000$} &  $0.950$\tiny{$\pm 0.001$} \\
\cline{1-12}
\cline{2-12}
\multirow{8}{*}{Robust} & \multirow{4}{*}{Yes} & 10  &  $0.562$\tiny{$\pm 0.006$} &  $0.567$\tiny{$\pm 0.008$} &  $0.697$\tiny{$\pm 0.006$} &  $0.661$\tiny{$\pm 0.003$} &  $0.708$\tiny{$\pm 0.002$} &  $0.147$\tiny{$\pm 0.005$} &  $0.971$\tiny{$\pm 0.000$} &  $0.953$\tiny{$\pm 0.000$} &  $0.940$\tiny{$\pm 0.005$} \\
       &    & 20  &  $0.562$\tiny{$\pm 0.006$} &  $0.567$\tiny{$\pm 0.008$} &  $0.696$\tiny{$\pm 0.006$} &  $0.660$\tiny{$\pm 0.003$} &  $0.706$\tiny{$\pm 0.001$} &  $0.132$\tiny{$\pm 0.007$} &  $0.971$\tiny{$\pm 0.000$} &  $0.953$\tiny{$\pm 0.000$} &  $0.940$\tiny{$\pm 0.005$} \\
       &    & 50  &  $0.562$\tiny{$\pm 0.006$} &  $0.567$\tiny{$\pm 0.009$} &  $0.696$\tiny{$\pm 0.006$} &  $0.656$\tiny{$\pm 0.003$} &  $0.705$\tiny{$\pm 0.003$} &  $0.135$\tiny{$\pm 0.006$} &  $0.971$\tiny{$\pm 0.000$} &  $0.953$\tiny{$\pm 0.000$} &  $0.940$\tiny{$\pm 0.005$} \\
       &    & 100 &  $0.562$\tiny{$\pm 0.006$} &  $0.567$\tiny{$\pm 0.008$} &  $0.696$\tiny{$\pm 0.006$} &  $0.653$\tiny{$\pm 0.001$} &  $0.703$\tiny{$\pm 0.003$} &  $0.129$\tiny{$\pm 0.005$} &  $0.971$\tiny{$\pm 0.000$} &  $0.953$\tiny{$\pm 0.000$} &  $0.940$\tiny{$\pm 0.005$} \\
\cline{2-12}
       & \multirow{4}{*}{No} & 10  &  $0.559$\tiny{$\pm 0.005$} &  $0.570$\tiny{$\pm 0.008$} &  $0.699$\tiny{$\pm 0.003$} &  $0.690$\tiny{$\pm 0.002$} &  $0.743$\tiny{$\pm 0.001$} &  $0.623$\tiny{$\pm 0.003$} &  $0.972$\tiny{$\pm 0.001$} &  $0.953$\tiny{$\pm 0.000$} &  $0.942$\tiny{$\pm 0.006$} \\
       &    & 20  &  $0.559$\tiny{$\pm 0.005$} &  $0.569$\tiny{$\pm 0.007$} &  $0.698$\tiny{$\pm 0.004$} &  $0.690$\tiny{$\pm 0.002$} &  $0.743$\tiny{$\pm 0.001$} &  $0.623$\tiny{$\pm 0.003$} &  $0.972$\tiny{$\pm 0.001$} &  $0.953$\tiny{$\pm 0.000$} &  $0.942$\tiny{$\pm 0.006$} \\
       &    & 50  &  $0.559$\tiny{$\pm 0.004$} &  $0.570$\tiny{$\pm 0.008$} &  $0.698$\tiny{$\pm 0.003$} &  $0.690$\tiny{$\pm 0.002$} &  $0.743$\tiny{$\pm 0.001$} &  $0.623$\tiny{$\pm 0.003$} &  $0.972$\tiny{$\pm 0.001$} &  $0.953$\tiny{$\pm 0.000$} &  $0.942$\tiny{$\pm 0.006$} \\
       &    & 100 &  $0.559$\tiny{$\pm 0.004$} &  $0.570$\tiny{$\pm 0.008$} &  $0.697$\tiny{$\pm 0.003$} &  $0.690$\tiny{$\pm 0.002$} &  $0.743$\tiny{$\pm 0.001$} &  $0.623$\tiny{$\pm 0.003$} &  $0.972$\tiny{$\pm 0.001$} &  $0.953$\tiny{$\pm 0.000$} &  $0.942$\tiny{$\pm 0.006$} \\
\bottomrule
\end{tabular}

\end{table*}

\paragraph{The attacker absolutely needs domain knowledge for successful attacks.}

\begin{figure*}
    \begin{subfigure}[b]{0.49\textwidth}
    \centering
    \includegraphics[width=\textwidth]{figures/url/acde_Standard_1.pdf}
    \caption{Standard target (A1, C1, D1, E1)}
    \end{subfigure}
    \begin{subfigure}[b]{0.49\textwidth}
    \centering
    \includegraphics[width=\textwidth]{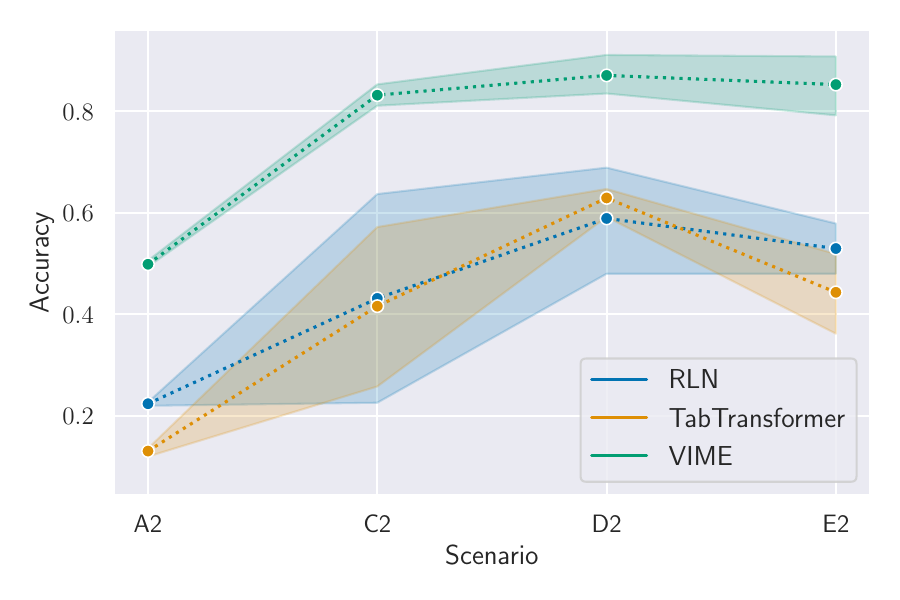}
    \caption{Standard target (A2, C2, D2, E2)}
    \end{subfigure}

    \begin{subfigure}[b]{0.49\textwidth}
    \centering
    \includegraphics[width=\textwidth]{figures/url/acde_Robust_1.pdf}
    \caption{Robust target (A1, C1, D1, E1)}
    \end{subfigure}
    \begin{subfigure}[b]{0.49\textwidth}
    \centering
    \includegraphics[width=\textwidth]{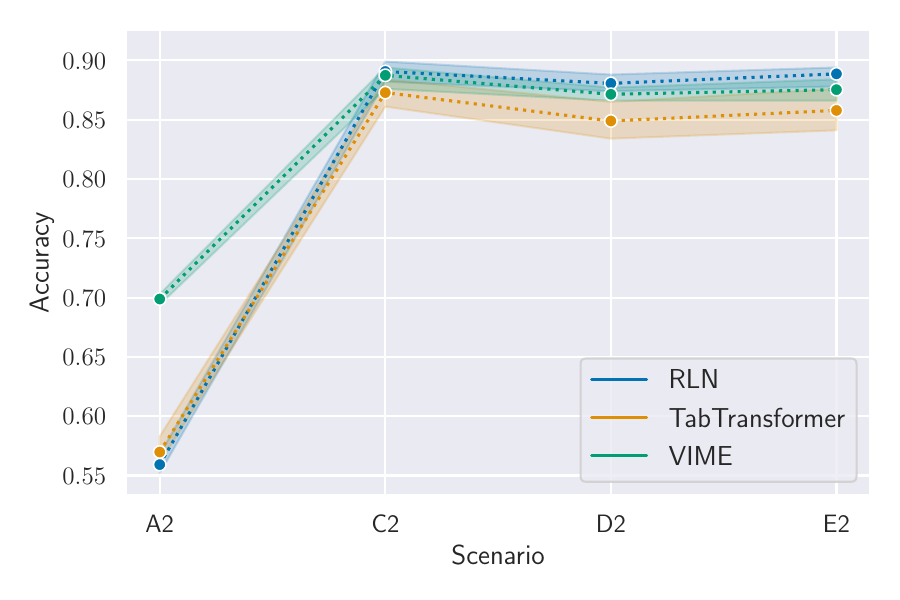}
    \caption{Robust target (A2, C2, D2, E2)}
    \end{subfigure}

    \caption{URL - Robust accuracy for different scenarios and target models. C1/C2 source is not robust. The range represents the 95\% confidence interval, over 5 seeds for A, and over 5 seeds and 2 source models (different from the target model) for C D E.}
    \label{fig:scenario_cde_url}
\end{figure*}

\begin{figure*}
    \begin{subfigure}[b]{0.49\textwidth}
    \centering
    \includegraphics[width=\textwidth]{figures/ctu_13_neris/acde_Standard_1.pdf}
    \caption{Standard target (A1, C1, D1, E1)}
    \end{subfigure}
    \begin{subfigure}[b]{0.49\textwidth}
    \centering
    \includegraphics[width=\textwidth]{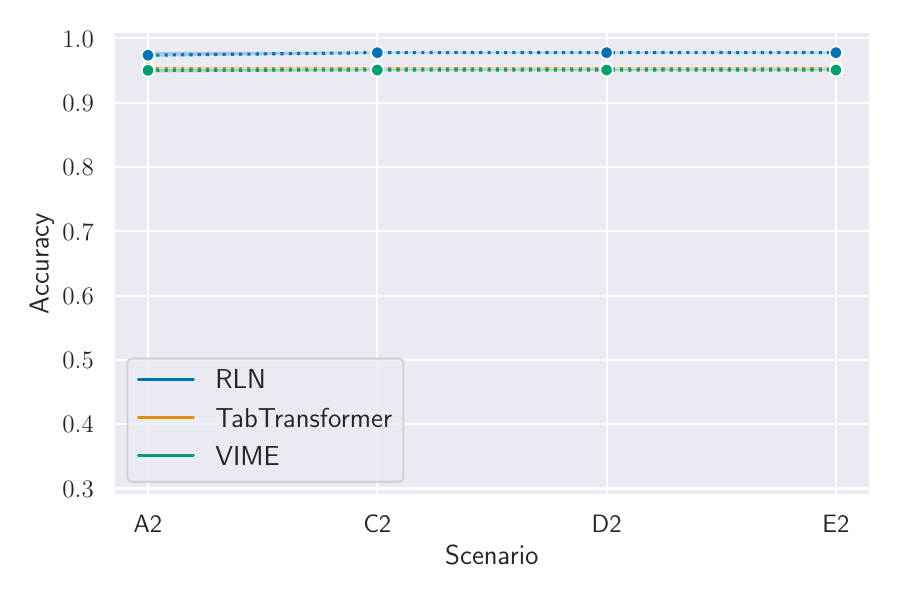}
    \caption{Standard target (A2, C2, D2, E2)}
    \end{subfigure}

    \begin{subfigure}[b]{0.49\textwidth}
    \centering
    \includegraphics[width=\textwidth]{figures/ctu_13_neris/acde_Robust_1.pdf}
    \caption{Robust target (A1, C1, D1, E1)}
    \end{subfigure}
    \begin{subfigure}[b]{0.49\textwidth}
    \centering
    \includegraphics[width=\textwidth]{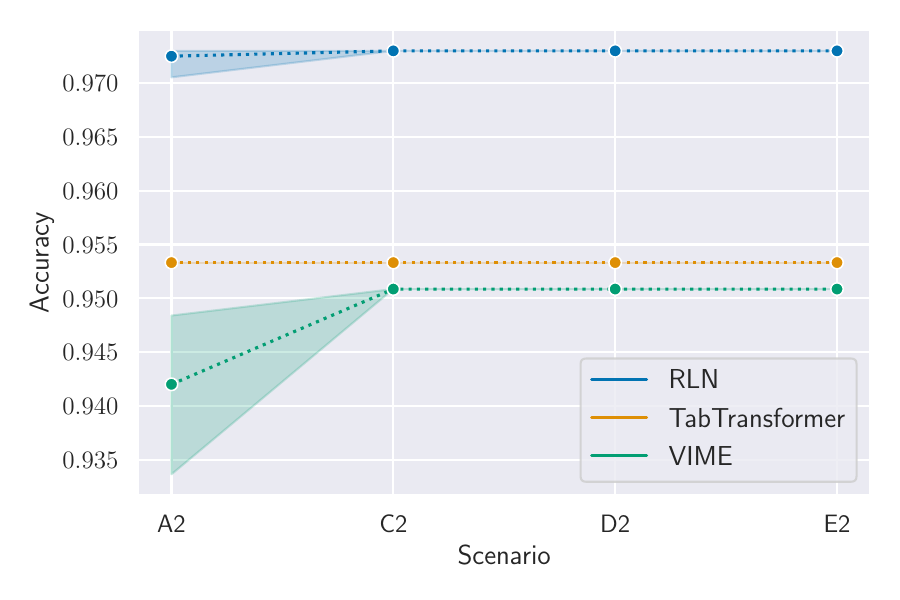}
    \caption{Robust target (A2, C2, D2, E2)}
    \end{subfigure}

    \caption{CTU - Robust accuracy for different scenarios and target models. C1/C2 source is not robust. The range represents the 95\% confidence interval, over 5 seeds for A, and over 5 seeds and 2 source models (different from the target model) for C D E.}
    \label{fig:scenario_cde_ctu}
\end{figure*}

We evaluated in the main paper the LCLD scenario, our conclusions stand for our remaining datasets URL (figure \ref{fig:scenario_cde_url}) and CTU (figure \ref{fig:scenario_cde_ctu}).

In table \ref{tab:acde_all}, we provide the detailed numerical results across our three datasets.

\begin{table*}
\centering
\caption{Numerical results of robust accuracy in the ACDE scenarios for each dataset.}
\footnotesize
\label{tab:acde_all}
\begin{tabular}{lll|lllllllll}
\toprule
    &        & Model & \multicolumn{3}{c}{RLN} & \multicolumn{3}{c}{TabTransformer} & \multicolumn{3}{c}{VIME} \\
    &        & {} &                   Mean\tiny{$\pm 0.95ci$} &    Min &    Max &                   Mean\tiny{$\pm 0.95ci$} &    Min &    Max &                   Mean\tiny{$\pm 0.95ci$} &    Min &    Max \\
Dataset & Training & Scenario &                            &        &        &                            &        &        &                            &        &        \\
\midrule
\multirow{16}{*}{URL} & \multirow{8}{*}{Standard} & A1 &  $0.123$\tiny{$\pm 0.006$} &  0.118 &  0.135 &  $0.088$\tiny{$\pm 0.002$} &  0.085 &  0.090 &  $0.492$\tiny{$\pm 0.005$} &  0.488 &  0.501 \\
    &        & A2 &  $0.224$\tiny{$\pm 0.003$} &  0.220 &  0.227 &  $0.131$\tiny{$\pm 0.006$} &  0.120 &  0.136 &  $0.499$\tiny{$\pm 0.004$} &  0.494 &  0.506 \\
    &        & C1 &  $0.397$\tiny{$\pm 0.148$} &  0.169 &  0.630 &  $0.357$\tiny{$\pm 0.134$} &  0.146 &  0.572 &  $0.840$\tiny{$\pm 0.004$} &  0.829 &  0.848 \\
    &        & C2 &  $0.431$\tiny{$\pm 0.129$} &  0.226 &  0.637 &  $0.416$\tiny{$\pm 0.098$} &  0.258 &  0.572 &  $0.832$\tiny{$\pm 0.010$} &  0.811 &  0.853 \\
    &        & D1 &  $0.533$\tiny{$\pm 0.034$} &  0.473 &  0.590 &  $0.563$\tiny{$\pm 0.023$} &  0.523 &  0.610 &  $0.885$\tiny{$\pm 0.021$} &  0.850 &  0.919 \\
    &        & D2 &  $0.589$\tiny{$\pm 0.062$} &  0.480 &  0.689 &  $0.629$\tiny{$\pm 0.010$} &  0.592 &  0.647 &  $0.871$\tiny{$\pm 0.022$} &  0.835 &  0.911 \\
    &        & E1 &  $0.518$\tiny{$\pm 0.036$} &  0.456 &  0.578 &  $0.405$\tiny{$\pm 0.073$} &  0.290 &  0.520 &  $0.861$\tiny{$\pm 0.034$} &  0.805 &  0.915 \\
    &        & E2 &  $0.530$\tiny{$\pm 0.028$} &  0.480 &  0.579 &  $0.444$\tiny{$\pm 0.046$} &  0.362 &  0.519 &  $0.853$\tiny{$\pm 0.035$} &  0.792 &  0.908 \\
\cline{2-12}
    & \multirow{8}{*}{Robust} & A1 &  $0.562$\tiny{$\pm 0.006$} &  0.551 &  0.569 &  $0.567$\tiny{$\pm 0.008$} &  0.554 &  0.575 &  $0.697$\tiny{$\pm 0.006$} &  0.687 &  0.703 \\
    &        & A2 &  $0.559$\tiny{$\pm 0.005$} &  0.552 &  0.568 &  $0.570$\tiny{$\pm 0.008$} &  0.561 &  0.583 &  $0.699$\tiny{$\pm 0.003$} &  0.694 &  0.703 \\
    &        & C1 &  $0.895$\tiny{$\pm 0.004$} &  0.887 &  0.905 &  $0.862$\tiny{$\pm 0.016$} &  0.835 &  0.889 &  $0.887$\tiny{$\pm 0.003$} &  0.881 &  0.893 \\
    &        & C2 &  $0.891$\tiny{$\pm 0.004$} &  0.882 &  0.899 &  $0.873$\tiny{$\pm 0.005$} &  0.861 &  0.884 &  $0.887$\tiny{$\pm 0.004$} &  0.876 &  0.894 \\
    &        & D1 &  $0.900$\tiny{$\pm 0.003$} &  0.892 &  0.908 &  $0.880$\tiny{$\pm 0.006$} &  0.867 &  0.893 &  $0.885$\tiny{$\pm 0.002$} &  0.882 &  0.890 \\
    &        & D2 &  $0.881$\tiny{$\pm 0.003$} &  0.874 &  0.888 &  $0.849$\tiny{$\pm 0.006$} &  0.834 &  0.865 &  $0.871$\tiny{$\pm 0.002$} &  0.866 &  0.877 \\
    &        & E1 &  $0.899$\tiny{$\pm 0.004$} &  0.890 &  0.908 &  $0.861$\tiny{$\pm 0.010$} &  0.845 &  0.878 &  $0.880$\tiny{$\pm 0.003$} &  0.873 &  0.884 \\
    &        & E2 &  $0.889$\tiny{$\pm 0.003$} &  0.878 &  0.894 &  $0.858$\tiny{$\pm 0.010$} &  0.841 &  0.876 &  $0.875$\tiny{$\pm 0.005$} &  0.866 &  0.884 \\
\cline{1-12}
\cline{2-12}
\multirow{16}{*}{LCLD} & \multirow{8}{*}{Standard} & A1 &  $0.001$\tiny{$\pm 0.001$} &  0.000 &  0.002 &  $0.135$\tiny{$\pm 0.006$} &  0.123 &  0.142 &  $0.066$\tiny{$\pm 0.005$} &  0.059 &  0.074 \\
    &        & A2 &  $0.542$\tiny{$\pm 0.008$} &  0.532 &  0.553 &  $0.671$\tiny{$\pm 0.002$} &  0.668 &  0.674 &  $0.655$\tiny{$\pm 0.002$} &  0.653 &  0.657 \\
    &        & C1 &  $0.185$\tiny{$\pm 0.014$} &  0.153 &  0.212 &  $0.231$\tiny{$\pm 0.019$} &  0.191 &  0.264 &  $0.282$\tiny{$\pm 0.004$} &  0.272 &  0.296 \\
    &        & C2 &  $0.673$\tiny{$\pm 0.004$} &  0.662 &  0.680 &  $0.686$\tiny{$\pm 0.011$} &  0.666 &  0.703 &  $0.661$\tiny{$\pm 0.001$} &  0.659 &  0.664 \\
    &        & D1 &  $0.309$\tiny{$\pm 0.050$} &  0.225 &  0.397 &  $0.226$\tiny{$\pm 0.007$} &  0.211 &  0.240 &  $0.366$\tiny{$\pm 0.100$} &  0.205 &  0.526 \\
    &        & D2 &  $0.679$\tiny{$\pm 0.002$} &  0.674 &  0.681 &  $0.700$\tiny{$\pm 0.001$} &  0.697 &  0.704 &  $0.666$\tiny{$\pm 0.000$} &  0.664 &  0.666 \\
    &        & E1 &  $0.231$\tiny{$\pm 0.012$} &  0.206 &  0.261 &  $0.200$\tiny{$\pm 0.012$} &  0.176 &  0.224 &  $0.289$\tiny{$\pm 0.066$} &  0.180 &  0.397 \\
    &        & E2 &  $0.680$\tiny{$\pm 0.001$} &  0.678 &  0.681 &  $0.691$\tiny{$\pm 0.008$} &  0.676 &  0.704 &  $0.663$\tiny{$\pm 0.002$} &  0.658 &  0.666 \\
\cline{2-12}
    & \multirow{8}{*}{Robust} & A1 &  $0.661$\tiny{$\pm 0.003$} &  0.659 &  0.666 &  $0.708$\tiny{$\pm 0.002$} &  0.706 &  0.711 &  $0.147$\tiny{$\pm 0.005$} &  0.138 &  0.151 \\
    &        & A2 &  $0.690$\tiny{$\pm 0.002$} &  0.686 &  0.692 &  $0.743$\tiny{$\pm 0.001$} &  0.742 &  0.744 &  $0.623$\tiny{$\pm 0.003$} &  0.619 &  0.626 \\
    &        & C1 &  $0.701$\tiny{$\pm 0.001$} &  0.699 &  0.702 &  $0.739$\tiny{$\pm 0.001$} &  0.737 &  0.741 &  $0.381$\tiny{$\pm 0.006$} &  0.361 &  0.395 \\
    &        & C2 &  $0.707$\tiny{$\pm 0.000$} &  0.707 &  0.707 &  $0.745$\tiny{$\pm 0.000$} &  0.744 &  0.745 &  $0.652$\tiny{$\pm 0.001$} &  0.649 &  0.653 \\
    &        & D1 &  $0.701$\tiny{$\pm 0.002$} &  0.695 &  0.705 &  $0.740$\tiny{$\pm 0.001$} &  0.737 &  0.742 &  $0.455$\tiny{$\pm 0.066$} &  0.349 &  0.565 \\
    &        & D2 &  $0.707$\tiny{$\pm 0.000$} &  0.706 &  0.707 &  $0.744$\tiny{$\pm 0.001$} &  0.743 &  0.745 &  $0.654$\tiny{$\pm 0.000$} &  0.653 &  0.654 \\
    &        & E1 &  $0.703$\tiny{$\pm 0.001$} &  0.700 &  0.706 &  $0.739$\tiny{$\pm 0.002$} &  0.735 &  0.742 &  $0.400$\tiny{$\pm 0.023$} &  0.358 &  0.446 \\
    &        & E2 &  $0.707$\tiny{$\pm 0.000$} &  0.707 &  0.707 &  $0.745$\tiny{$\pm 0.000$} &  0.745 &  0.745 &  $0.653$\tiny{$\pm 0.000$} &  0.652 &  0.654 \\
\cline{1-12}
\cline{2-12}
\multirow{16}{*}{CTU} & \multirow{8}{*}{Standard} & A1 &  $0.940$\tiny{$\pm 0.002$} &  0.936 &  0.943 &  $0.953$\tiny{$\pm 0.000$} &  0.953 &  0.953 &  $0.408$\tiny{$\pm 0.047$} &  0.322 &  0.450 \\
    &        & A2 &  $0.974$\tiny{$\pm 0.002$} &  0.973 &  0.978 &  $0.953$\tiny{$\pm 0.000$} &  0.953 &  0.953 &  $0.950$\tiny{$\pm 0.001$} &  0.948 &  0.951 \\
    &        & C1 &  $0.978$\tiny{$\pm 0.000$} &  0.978 &  0.978 &  $0.953$\tiny{$\pm 0.000$} &  0.953 &  0.953 &  $0.951$\tiny{$\pm 0.000$} &  0.951 &  0.951 \\
    &        & C2 &  $0.978$\tiny{$\pm 0.000$} &  0.978 &  0.978 &  $0.953$\tiny{$\pm 0.000$} &  0.953 &  0.953 &  $0.951$\tiny{$\pm 0.000$} &  0.951 &  0.951 \\
    &        & D1 &  $0.978$\tiny{$\pm 0.000$} &  0.978 &  0.978 &  $0.953$\tiny{$\pm 0.000$} &  0.953 &  0.953 &  $0.951$\tiny{$\pm 0.000$} &  0.951 &  0.951 \\
    &        & D2 &  $0.978$\tiny{$\pm 0.000$} &  0.978 &  0.978 &  $0.953$\tiny{$\pm 0.000$} &  0.953 &  0.953 &  $0.951$\tiny{$\pm 0.000$} &  0.951 &  0.951 \\
    &        & E1 &  $0.978$\tiny{$\pm 0.000$} &  0.978 &  0.978 &  $0.953$\tiny{$\pm 0.000$} &  0.953 &  0.953 &  $0.951$\tiny{$\pm 0.000$} &  0.951 &  0.951 \\
    &        & E2 &  $0.978$\tiny{$\pm 0.000$} &  0.978 &  0.978 &  $0.953$\tiny{$\pm 0.000$} &  0.953 &  0.953 &  $0.951$\tiny{$\pm 0.000$} &  0.951 &  0.951 \\
\cline{2-12}
    & \multirow{8}{*}{Robust} & A1 &  $0.971$\tiny{$\pm 0.000$} &  0.971 &  0.971 &  $0.953$\tiny{$\pm 0.000$} &  0.953 &  0.953 &  $0.940$\tiny{$\pm 0.005$} &  0.934 &  0.948 \\
    &        & A2 &  $0.972$\tiny{$\pm 0.001$} &  0.971 &  0.973 &  $0.953$\tiny{$\pm 0.000$} &  0.953 &  0.953 &  $0.942$\tiny{$\pm 0.006$} &  0.934 &  0.948 \\
    &        & C1 &  $0.973$\tiny{$\pm 0.000$} &  0.973 &  0.973 &  $0.953$\tiny{$\pm 0.000$} &  0.953 &  0.953 &  $0.951$\tiny{$\pm 0.000$} &  0.951 &  0.951 \\
    &        & C2 &  $0.973$\tiny{$\pm 0.000$} &  0.973 &  0.973 &  $0.953$\tiny{$\pm 0.000$} &  0.953 &  0.953 &  $0.951$\tiny{$\pm 0.000$} &  0.951 &  0.951 \\
    &        & D1 &  $0.973$\tiny{$\pm 0.000$} &  0.973 &  0.973 &  $0.953$\tiny{$\pm 0.000$} &  0.953 &  0.953 &  $0.951$\tiny{$\pm 0.000$} &  0.951 &  0.951 \\
    &        & D2 &  $0.973$\tiny{$\pm 0.000$} &  0.973 &  0.973 &  $0.953$\tiny{$\pm 0.000$} &  0.953 &  0.953 &  $0.951$\tiny{$\pm 0.000$} &  0.951 &  0.951 \\
    &        & E1 &  $0.973$\tiny{$\pm 0.000$} &  0.973 &  0.973 &  $0.953$\tiny{$\pm 0.000$} &  0.953 &  0.953 &  $0.951$\tiny{$\pm 0.000$} &  0.951 &  0.951 \\
    &        & E2 &  $0.973$\tiny{$\pm 0.000$} &  0.973 &  0.973 &  $0.953$\tiny{$\pm 0.000$} &  0.953 &  0.953 &  $0.951$\tiny{$\pm 0.000$} &  0.951 &  0.951 \\
\bottomrule
\end{tabular}
\end{table*}

\end{document}